\RequirePackage{fix-cm}
\documentclass{article}

\usepackage{iclr2027_conference,times}
\usepackage[T1]{fontenc}
\usepackage[section]{placeins}
\usepackage{needspace}
\usepackage{microtype}
\usepackage{amsmath,amssymb}
\usepackage{booktabs}
\usepackage{multirow}
\usepackage{array}
\usepackage{tabularx}
\usepackage{longtable}
\usepackage{enumitem}
\usepackage{graphicx}
\usepackage{xcolor}
\usepackage{caption}
\usepackage{hyperref}
\usepackage{url}

\hypersetup{
  colorlinks=true,
  linkcolor=blue!50!black,
  citecolor=blue!50!black,
  urlcolor=blue!60!black,
  pdftitle={ActiveArena: Benchmarking and Understanding Active Perception in Robotic Manipulation},
  pdfauthor={Yibo Li, Enshen Zhou, Rui Chen, Yanjun Ding, Mengzhen Liu, Yi Han, Jiabo Zhan, Lipeng Wang, Shanghang Zhang, Lu Sheng}
}

\setlist{nosep}
\renewcommand{\arraystretch}{1.12}

\newenvironment{reporttable}[1][\linewidth]{%
  \par\addvspace{10pt plus 2pt minus 2pt}%
  \begingroup\centering\noindent\begin{minipage}{#1}%
  \centering\captionsetup{type=table,position=top}%
}{%
  \end{minipage}\par\endgroup\addvspace{10pt plus 2pt minus 2pt}%
}

\newcommand{\mnm}{{\fontshape{n}\selectfont\itshape ActiveArena}}
\newcommand{\vs}{}
\newcommand{\ie}{\emph{i.e.}}
\newcommand{\eg}{\emph{e.g.}}

\newcommand{\equalmark}{\textsuperscript{*}}
\newcommand{\correspondingmark}{\textsuperscript{\ensuremath{\dagger}}}

\title{ActiveArena: Benchmarking and\\
Understanding Active Perception in\\
Robotic Manipulation}
\author{%
  \normalsize\bfseries
  Yibo Li\textsuperscript{1,2}\equalmark,\enspace
  Enshen Zhou\textsuperscript{1,2}\equalmark,\enspace
  Rui Chen\textsuperscript{1},\enspace
  Yanjun Ding\textsuperscript{1},\enspace
  Mengzhen Liu\textsuperscript{2,3}\\[2pt]
  \normalsize\bfseries Yi Han\textsuperscript{1,2},\enspace
  Jiabo Zhan\textsuperscript{4},\enspace
  Lipeng Wang\textsuperscript{1,2},\enspace
  Shanghang Zhang\textsuperscript{2,3},\enspace
  Lu Sheng\textsuperscript{1,2}\correspondingmark\\[6pt]
  \normalfont\small \textsuperscript{1}Beihang University\quad
  \textsuperscript{2}Beijing Academy of Artificial Intelligence\\
  \normalfont\small \textsuperscript{3}Peking University\quad
  \textsuperscript{4}Tsinghua University\\[4pt]
  {\normalfont\scriptsize
    \equalmark\ Equal contribution.\quad
    \correspondingmark\ Corresponding author.}
}
\date{}

\iclrfinalcopy
\begin{document}
\maketitle
\lhead{}

\begin{abstract}

Active perception and manipulation are crucial for robots to interact with complex scenes.
Existing benchmarks struggle to evaluate how robots effectively acquire and maintain information in memory in an active manner. 
To this end, we introduce \mnm-Sim, an active-perception simulator with controllable viewpoints and large-scale workspaces as the foundation.
Built on this, we propose \mnm-Bench, which comprises 35 tasks across 5 fine-grained categories, covering visual exploration and interactive information acquisition.
Each task is difficult to solve from passive observations alone, requiring multi-round evidence acquisition and memory-based reasoning. 
The benchmark provides rich memory annotations, standardized training data, and ID/OOD protocols featuring disjoint scenes, unseen distractor configurations, and novel backgrounds.
Moreover, we present \mnm-VLA, a modular suite of 13 vision--language--action configurations for controlled studies of memory writing, memory capacity, proprioceptive state, subtask supervision, and high-level planning in active perception.
Benchmark results reveal a substantial ID--OOD gap: uniform memory sampling, increased memory capacity under reliable write policies, proprioceptive inputs, and subtask supervision improve OOD generalization, while planner-guided memory management and decision-making achieve performance close to the best-performing configuration using only sparse memory.
{\mnm} thus provides a unified testbed to develop and diagnose models for active perception and manipulation.
\end{abstract}

\noindent Our code and data are open-source: \url{https://leeibo.github.io/ActiveArena}.

% Main text.
\section{Introduction}
% Active perception is fundamental to robotic autonomy in complex, open environments. It couples two complementary processes: acting to perceive, whereby a robot deliberately changes its viewpoint or interacts with the environment to acquire missing task-relevant evidence, and perceiving to act, whereby it assesses whether the available evidence is sufficient and uses accumulated observations to determine subsequent actions. Together, they form a closed perception--action loop in which the robot gathers evidence, integrates information across time and viewpoints, and decides whether to manipulate or continue sensing.

Active perception is crucial for embodied AI, as robot agents should actively perceive unstructured scenes and act accordingly, like a human~\citep{bajcsy1988active}.
This requires two complementary abilities: \textbf{(1)} maintaining a memory of acquired and missing information, and \textbf{(2)} deciding what information to acquire next and how to obtain it via a perception-action loop~\citep{yang2025thinking,wang2506mindcube,xiong2025vision,liu2026activevla,liu2026sapave,hong2026esibench,he2026towards,li2026act}.
Specifically, consider finding a specific novel hidden in a backpack within a cluttered workspace. 
The robot alternates between viewpoint adjustment (\eg, resolve occlusions) and exploratory interaction (\eg, open containers, search backpack) in a closed loop.
It then assesses whether the evidence in memory is sufficient (\eg, verify retrieved book) and uses its memory to guide subsequent action (\eg, search unexplored regions).
Thus, a comprehensive evaluation of active perception should jointly measure \textit{active information acquisition} and \textit{memory-based information maintenance}, as either dimension alone provides only a partial view of such ability.

% Fig.~\ref{fig:active_perception_rollout} shows representative closed-loop rollouts for visual search and interactive information acquisition. The robot alternates between observing, grounding evidence in the current view and memory, acquiring missing information through search or interaction, acting, and updating its memory.

\begin{figure}[!tp]

    \centering
    
    \includegraphics[width=\textwidth]{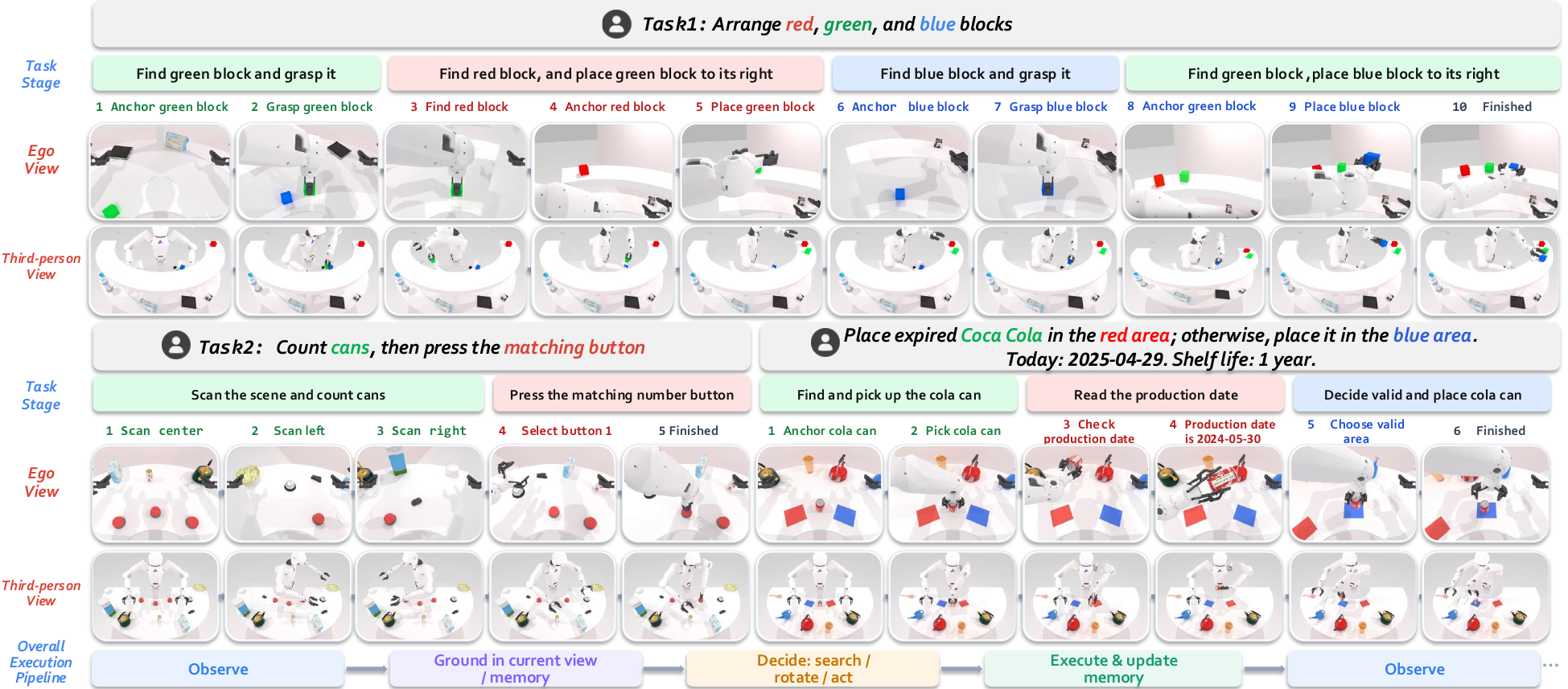}
    % \vspace{-4mm}
    \caption{Representative {\mnm} rollouts covering 2 families (\ie, visual exploration, interactive acquisition). Each example shows the perception-action loop of evidence grounding, information acquisition, action execution, and memory update by coordinated head/arm actions in a large workspace. The third-person view is included solely for scene visualization.}
    \label{fig:active_perception_rollout}
    \vs
\end{figure}

% Benchmarks for visual search, environment exploration, or high-level planning often decouple decision making from continuous robot control.

% Conversely, conventional manipulation benchmarks typically provide nearly complete task information from fixed cameras, making active information acquisition unnecessary.

% Fair comparison is further hindered by incompatible memory representations and interaction interfaces across existing methods. 

% Current simulators are also commonly restricted to fixed viewpoints, small workspaces, or limited support for revealing hidden information through physical interaction. 

% Consequently, there remains no unified testbed that jointly evaluates information acquisition, spatiotemporal memory, and low-level control.

% Despite its importance, active perception and manipulation remain poorly captured by existing benchmarks.

Recent Vision-Language-Action (VLA) and World-Action Models (WAM) benchmarks have brought memory evaluation in long-horizon embodied tasks to the forefront~\citep{cherepanov2025mikasa,fang2025sam2act,lei2026robomemarena,dai2026robomme,chen2026rmbench,shah2026prism}.
However, existing benchmarks overlook two key dimensions of active perception:
\textbf{(1)} \textit{Active information acquisition}: they assume fixed-view observations are sufficient and do not require robots to seek missing or out-of-view evidence.
\textbf{(2)} \textit{Active information maintenance}: they assume complete information in memory, neglecting the selective preservation and updating of information as observations accumulate.
These omissions stem from a fundamental limitation of current simulators: fixed viewpoints create an \underline{embodiment gap}, confined workspaces a \underline{scene gap}, and limited support for acquiring information an \underline{interaction gap}.
In other words, \textit{despite the importance of active perception, neither a unified simulator addressing all three gaps nor a benchmark systematically evaluating both core dimensions has been explored}.

To this end, we present \textbf{ActiveArena}, the first comprehensive and standardized simulator--benchmark--baseline suite to meet above expectation.
At its core, \textbf{ActiveArena-Sim}, is designed at three levels:
\textbf{(1)} \textit{Embodiment}: controllable head and torso joints enable active viewpoint selection, providing 6.3x greater visible-scene coverage than conventional fixed-view simulators.
\textbf{(2)} \textit{Scene}: a $180^\circ$ multi-level workspace 2x the interaction area and distributes task-relevant objects across diverse directions and heights.
\textbf{(3)} \textit{Interaction}: robots actively reveal hidden evidence by manipulating objects and inspecting them from informative viewpoints, enabling richer information acquisition (2x) than passive observation.
This simulator forms the foundation of the entire suite.

Built on \textbf{ActiveArena-Sim}, \textbf{ActiveArena-Bench} comprises tasks that are nearly impossible to solve from passive observations alone.
It spans two complementary families---visual exploration and interactive acquisition---depending on how task-relevant evidence should be actively revealed.
The benchmark further organizes tasks into five categories based on the number of relevant objects and perception--action rounds (up to 9), and jointly assesses active spatial search, cross-view memory, interactive evidence acquisition, and multi-round closed-loop reasoning. 
It also provides multi-granularity memory annotations to support diverse memory and model design for deep analysis.

As real-world active perception and manipulation demand robust generalization, we establish a rigorous, standardized training and evaluation protocol that strictly separates in-distribution (ID) and out-of-distribution (OOD) settings.
The ID setting evaluates the same domain task learning, while the OOD setting introduces unseen distractor layouts and backgrounds to test whether models generalize information-acquisition and memory-maintenance strategies rather than exploit spurious environmental regularities.
Our results expose a critical gap: different memory and model designs that achieve strong ID performance do not necessarily generalize OOD robustly in active perception and manipulation.

% To disentangle in-distribution task learning from active-perception generalization, we establish a standardized training and evaluation protocol. We provide training data containing the fields required by the evaluated strategies and assess all methods under complementary in-distribution (ID) and out-of-distribution (OOD) settings. The ID setting measures task learning within the training distribution, whereas the OOD setting introduces unseen distractor configurations and environmental backgrounds. This design tests whether a policy can generalize its information-acquisition and decision-making strategies beyond the training distribution, rather than exploit fixed environmental regularities. Our results reveal a substantial gap: strong ID performance does not necessarily translate into robust OOD generalization.

We further introduce \textbf{ActiveArena-VLA}, a modular evaluation suite spanning 13 VLA configurations for controlled analysis of memory writing, memory capacity, proprioception, subtask supervision, and high-level planning.
Our results reveal that: 
\textbf{(1)} Uniform temporal sampling generalizes better OOD than event-triggered writing based on subtask transitions or robot motion.
\textbf{(2)} Larger memory is not inherently beneficial; its effectiveness depends critically on write-policy stability.
\textbf{(3)} Proprioceptive inputs and auxiliary subtask supervision provide additional gains. 
\textbf{(4)} Explicit planner-mediated memory management and decision-making achieve performance close to our best-performing \mnm-VLA variant while using only sparse memory.
Given that these designs still overcome the ID-OOD gap in active perception, these findings highlight the long-term value of research in this domain.

Our main contributions are threefold:
\textbf{(1)} We introduce \textbf{ActiveArena-Sim}, a simulator designed for active viewpoint control, large-workspace manipulation, and spatio-temporal memory research.
\textbf{(2)} We develop \textbf{ActiveArena-Bench}, a unified benchmark of 35 tasks spanning 2 task families and 5 categories, with rich memory annotations, standardized training data, and ID/OOD protocols for evaluating task learning and active-perception generalization.
\textbf{(3)} We present \textbf{ActiveArena-VLA}, a modular suite of 13 VLA variants, and systematically study how memory writing, memory capacity, proprioception, subtask supervision, and high-level planning shape active-perception performance and generalization.

% \end{enumerate}
% Required in the preamble:
% \usepackage{amssymb,graphicx,tikz}
\section{\mnm}
\label{ch:ch3}

\begin{figure}[!tp]

    \centering
    \includegraphics[width=\textwidth]{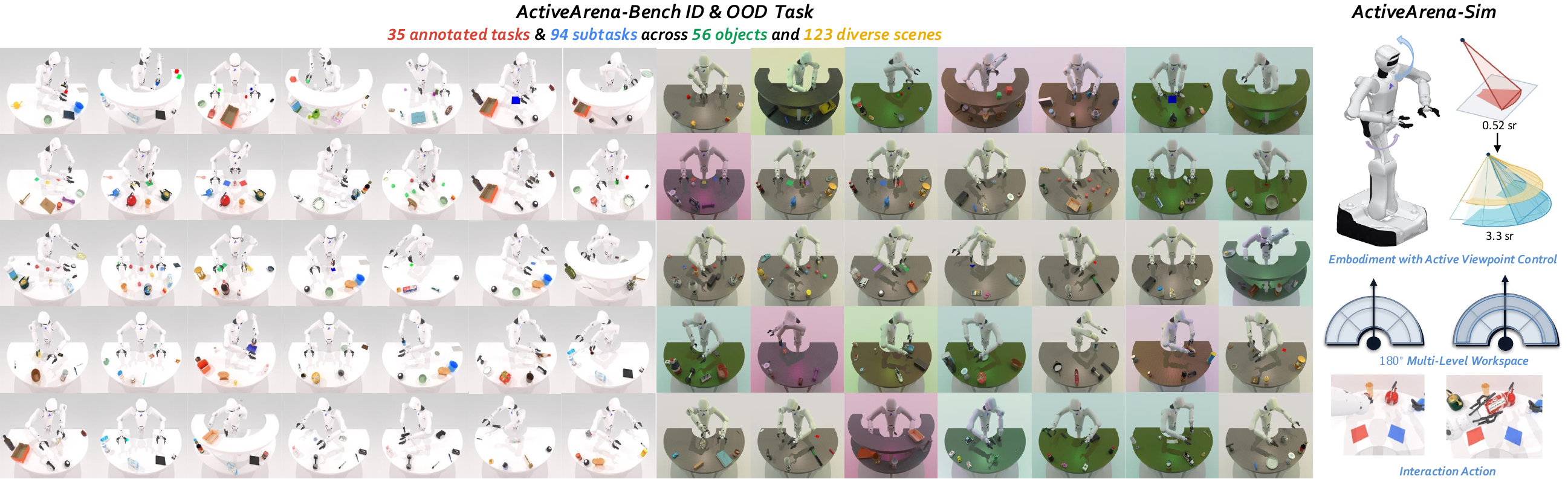}
    
    \caption{Overview of \mnm-Bench and \mnm-Sim. The third-person view is included solely for scene visualization.}
    \label{fig:dataset_overview}
    
\end{figure}

This section presents \mnm, a simulator-benchmark-baseline suite for active perception and manipulation.
We first formulate active-perception manipulation under partial observability.
We then elaborate on the design of \mnm-Sim.
Finally, we introduce task-design principles and taxonomy of the \mnm-Bench.
The position of \mnm-Bench among related benchmarks is summarized in Table~\ref{tab:benchmark_comparison}.

\noindent\textbf{Problem Formulation.}
  We consider language-conditioned robotic manipulation under partial
  observability. At time step $t$, the policy receives a visual observation
  $o_t$, an optional proprioceptive state $x_t$, and a language instruction
  $l$. The preceding interaction history is
  \begin{equation}
  h_t =
  \left(
  (o_\tau, x_\tau, \mathbf{a}_\tau)
  \right)_{\tau=0}^{t-1},
  \label{eq:interaction_history}
  \end{equation}
  where $x_\tau$ is omitted when proprioception is unavailable. A
  method-dependent memory summarizes this history as
  \begin{equation}
  M_t = \mathcal{M}_{\phi}(h_t),
  \qquad
  \mathcal{I}_t = (o_t, x_t, M_t, l).
  \label{eq:active_information_state}
  \end{equation}

\noindent Conceptually, the policy may take an action with one of two functional roles:
  acquiring task-relevant information or directly advancing task completion. Let
  $z_t \in \{\mathrm{info}, \mathrm{task}\}$ denote this role. The next action is
  then represented as
  \begin{equation}
  \mathbf{a}_t \sim
  \begin{cases}
  \pi_l^{\mathrm{info}}
  \left(\cdot \mid \mathcal{I}_t\right),
  & z_t = \mathrm{info}, \\[2pt]
  \pi_l^{\mathrm{task}}
  \left(\cdot \mid \mathcal{I}_t\right),
  & z_t = \mathrm{task}.
  \end{cases}
  \label{eq:active_perception_policy}
  \end{equation}
  Here, $\pi_l^{\mathrm{info}}$ covers actions that expose or localize missing
  evidence, including viewpoint adjustment, spatial exploration, container
  opening, and object reorientation. In contrast,
  $\pi_l^{\mathrm{task}}$ uses the accumulated evidence to execute the requested
  manipulation. The policy need not explicitly predict $z_t$; it denotes the
  functional role of the selected action.

  After executing $\mathbf{a}_t$, the policy receives $o_{t+1}$ and updates its
  memory to $M_{t+1}$. Because the same motor primitive may acquire information
  in one context and advance task completion in another, the two roles are
  distinguished by their task-level function rather than by disjoint action
  spaces. Active perception therefore requires the policy to alternate between
  information acquisition and task execution according to the evidence retained
  in memory.

\newcommand{\cmark}{%
  \textcolor{green!50!black}{\ensuremath{\checkmark}}%
}
\newcommand{\xmark}{%
  \textcolor{red!80!black}{\ensuremath{\times}}%
}
\begin{table}[!htb]
\caption{ Comparison of benchmarks across 7 dimensions. \textbf{Dynamic Head Viewpoint(DV)}: independently controllable viewpoint changes (excluding motion induced solely by the arm or wrist); \textbf{Bimanual(BM)}: bimanual manipulation; \textbf{Active Perception(AP)}: task-relevant information not initially visible; \textbf{Manipulation Action Output(AO)}: executable manipulation-action output; \textbf{Subtask Annotation(SA)} : temporally aligned subtask annotations; \textbf{Native Diverse Keyframes(NK)}: native semantic, stage, or event keyframes; \textbf{Real-World Counterpart(RW)}: equipped with a corresponding real-robot task, dataset, experiment, or evaluation. \cmark/\xmark indicate support or not.}
\label{tab:benchmark_comparison}
  \centering

    \fontsize{8.5}{10.5}\selectfont
    \setlength{\tabcolsep}{1.2pt}
    \renewcommand{\arraystretch}{1.05}

    \begin{tabular*}{\columnwidth}{
      @{\extracolsep{\fill}}
      l*{7}{c}
      @{}
    }
      \toprule
      \textbf{Benchmark}
      & \textbf{DV}
      & \textbf{BM}
      & \textbf{AP}
      & \textbf{AO}
      & \textbf{SA}
      & \textbf{NK}
      & \textbf{RW} \\
      \midrule

      EmbodiedEval~\citep{cheng2025embodiedeval}
      & \cmark & \xmark & \cmark & \xmark
      & \xmark & \xmark & \xmark \\

      CHAIN~\citep{wu2026chain}
      & \xmark & \xmark & \cmark & \xmark
      & \xmark & \xmark & \xmark \\

      ESI-Bench~\citep{hong2026esibench}
      & \cmark & \xmark & \cmark & \xmark
      & \xmark & \xmark & \xmark \\

      RoboTwin 2.0~\citep{chen2025robotwin2}
      & \xmark & \cmark & \xmark & \cmark
      & \xmark & \xmark & \cmark \\

      RoboDojo~\citep{chen2026robodojo}
      & \xmark & \cmark & \xmark & \cmark
      & \xmark & \xmark & \cmark \\

      LIBERO~\citep{liu2023libero}
      & \xmark & \xmark & \xmark & \cmark
      & \xmark & \xmark & \xmark \\

      LIBERO-Plus~\citep{fei2025liberoplus}
      & \xmark & \xmark & \xmark & \cmark
      & \xmark & \xmark & \xmark \\

      MIKASA~\citep{cherepanov2025mikasa}
      & \xmark & \xmark & \xmark & \cmark
      & \xmark & \xmark & \cmark \\

      RoboMemArena~\citep{lei2026robomemarena}
      & \xmark & \xmark & \xmark & \cmark
      & \cmark & \cmark & \cmark \\

      RoboMME~\citep{dai2026robomme}
      & \xmark & \xmark & \xmark & \cmark
      & \cmark & \cmark & \cmark \\

      RMBench~\citep{chen2026rmbench}
      & \xmark & \cmark & \xmark & \cmark
      & \cmark & \xmark & \cmark \\

      MemoryBench~\citep{fang2025sam2act}
      & \xmark & \xmark & \xmark & \cmark
      & \xmark & \xmark & \cmark \\

      ReMemBench~\citep{shah2026prism}
      & \cmark & \xmark & \xmark & \cmark
      & \xmark & \xmark & \cmark \\

      RoboCasa365~\citep{nasiriany2026robocasa365}
      & \cmark & \xmark & \xmark & \cmark
      & \cmark & \xmark & \cmark \\

     EFM-10~\citep{he2026towards}
      & \xmark & \cmark & \cmark & \cmark
      & \xmark & \xmark & \cmark \\

      \midrule
      \textbf{\mnm-Bench}
      & \cmark & \cmark & \cmark & \cmark
      & \cmark & \cmark & \cmark \\

      \bottomrule
    \end{tabular*}

      \vs
      % \vspace{-2mm}
\end{table}

\begin{figure}[!tp]
\vs
    \centering
    \includegraphics[width=\textwidth]{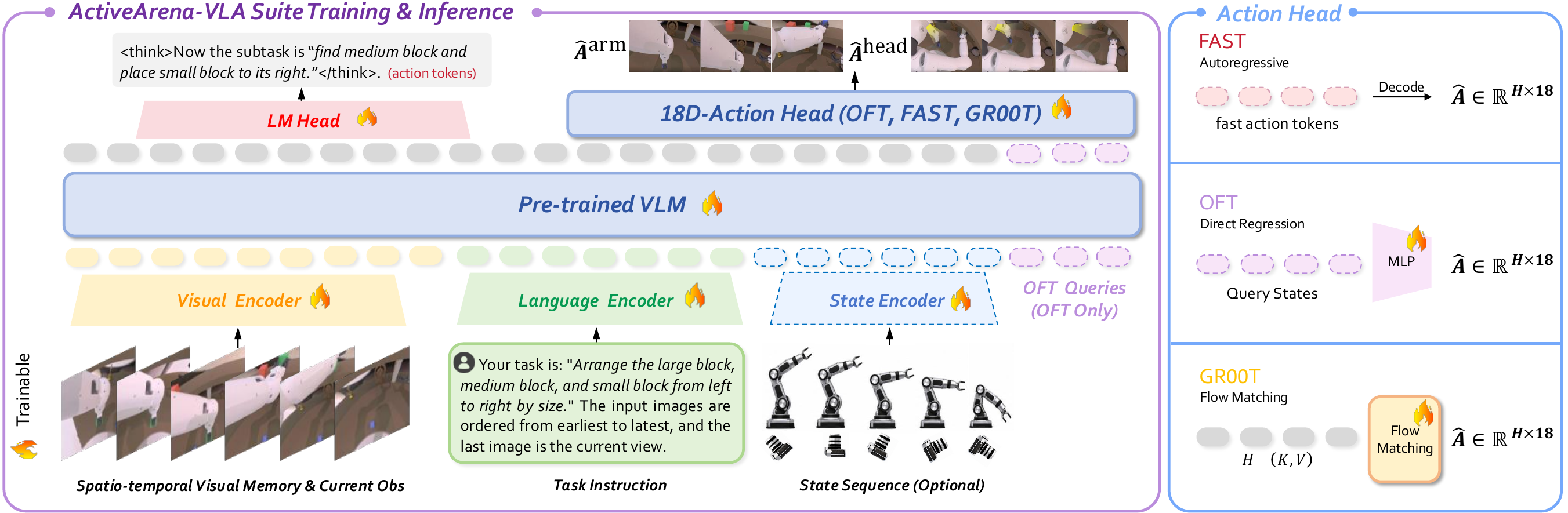}
   \caption{Overview of the \mnm-VLA framework. Spatio-temporal visual memory, language, and optional proprioceptive state sequences are encoded by a VLM and connected to OFT, FAST, or GR00T action heads for 18-D action prediction.}
    \label{fig:activearena_vla_suite}
    
\end{figure}

\noindent\textbf{Simulation Workspace Design.}
Upon RoboTwin~2.0~\citep{chen2025robotwin2}, we redesign its workspace to make active information acquisition necessary.
In the original setup, manipulation tasks are concentrated on a local tabletop directly in front of the robot, and task-relevant objects are observable from a fixed camera.
Although suitable for evaluating basic manipulation skills, this configuration provides limited support for tasks that genuinely require active perception.
We replace the tabletop with a workspace whose horizontal footprint spans an approximately $180^\circ$ sector centered on the robot, and parameterize object placement using robot-centric cylindrical coordinates.
No single fixed view covers the entire scene, requiring the robot to rotate its head or torso to search for and revisit task-relevant objects.
Such viewpoint changes frequently move previously observed objects outside the current field of view, making cross-view memory essential.
For selected tasks, we further introduce a two-tier layout that distributes objects across different heights and spatial regions that cannot be observed simultaneously.
The resulting workspace is shown in Fig.~\ref{fig:dataset_overview}.

\noindent\textbf{Embodiment with Active Viewpoint Control.}
The default embodiments in RoboTwin~2.0 use fixed-camera configurations and are therefore unsuitable for systematically evaluating active visual search and cross-view manipulation.
We integrate the Astribot S1~\citep{gao2025astribot} dual-arm humanoid robot into the simulator and enable control over its head and torso degrees of freedom, allowing the robot to actively adjust its viewpoint.
The resulting embodiment is shown in Fig.~\ref{fig:dataset_overview}.
The robot action space contains controls for both arms, both grippers, torso, and  head, yielding an 18-D action vector:
\begin{equation}
\mathbf{a}_t =
\left[
\mathbf{a}^{L}_t,
g^{L}_t,
\mathbf{a}^{R}_t,
g^{R}_t,
a^{\mathrm{torso}}_t,
a^{\mathrm{head}}_t
\right]
\in \mathbb{R}^{18}.
\end{equation}

\noindent\textbf{Process-Level Annotations.} We extend trajectory logging with process-level annotations for active perception. Each trajectory is decomposed into semantic subtasks and three functional phases: \emph{search}, \emph{anchor}, and \emph{action}. Search acquires missing evidence, anchor establishes or recovers an evidence-bearing view, and action executes manipulation or reveals additional information. Annotations are aligned frame-wise with observations, robot states, and actions. They include target visibility, discovery history, image-space location, camera orientation, and manipulation targets. Visibility is computed from the visible projected area of each 3D bounding box to account for occlusion and image truncation. For cross-view tasks, informative keyframes and their viewing directions remain annotated after the target leaves the current view. For interaction-dependent tasks, such as inspecting the production date on the back of a can, information-revealing actions and inspection views are linked to the decisions they support. These annotations enable training and evaluation of visual search, evidence localization, keyframe selection, memory retrieval, viewpoint recovery, and action prediction without assuming a specific memory architecture.

\noindent\textbf{Task-Design Principles.}
Every task satisfies 4 requirements:
\textbf{(1)} \textit{Initial Information Insufficiency.}
The correct manipulation cannot be determined from the initial observation alone.
\textbf{(2)} \textit{Information Accessibility.}
The missing task-relevant information can be acquired through spatial search or physical interaction.
\textbf{(3)} \textit{Historical Dependence.}
Evidence acquired from previous observations must remain relevant even after the corresponding objects or regions leave the current field of view.
\textbf{(4)} \textit{Closed-Loop Decision Making.}
Newly acquired evidence must influence subsequent localization or manipulation decisions.
After each action, the robot must reassess whether the accumulated evidence is sufficient and continue acquiring information when necessary.

\noindent\textbf{Task Taxonomy.}
\label{ch3: task taxonomy}
Following these principles, we construct 35 simulated manipulation tasks.
At the coarsest level, the tasks are divided into two families: Visual Search and Interactive Information Acquisition.
Visual-search tasks are further organized according to the number of task-relevant objects and the number of required perception--action loops, resulting in five categories overall:
\textbf{(1)} \textit{Single-Object Search (SS).}
Locate a single target and immediately perform the required manipulation.
These tasks require the robot to use its limited memory to identify unexplored regions.
\textbf{(2)} \textit{Single-Object Loop (SL).}
Complete cross-region search and the transport of a single object.
\textbf{(3)} \textit{Multi-Object Decision (MD).} Search among multiple candidates and select the correct target or action from the acquired evidence.
\textbf{(4)} \textit{Multi-Object Loop (ML).} Complete multiple perception--action loops across objects and spatial regions.
\textbf{(5)} \textit{Interactive Information Acquisition (IA).}
The robot must physically interact with the environment to reveal otherwise hidden task-relevant information before determining the appropriate manipulation.
Representative task rollouts are provided in Fig.~\ref{fig:active_perception_rollout}.

\section{Experiments}

\textbf{Data and Evaluation Protocol.}
Evaluation confined to the training scene distribution can conflate in-distribution imitation with genuine policy generalization. 
To disentangle imitation learning from generalization, we introduce two evaluation settings: ID and OOD. 
For benchmark adaptation, every method is trained on the same set of 100 ID trajectories per task. Across 35 tasks, the training set contains a total of 581.2k frames, corresponding to 10.76 hours of video. Each model is trained for 3.9 epochs while consuming only the annotation fields required by its architecture. During evaluation, all models are tested using the same set of 50 randomly generated seeds for each of the ID and OOD settings.
ID scenes contain randomized task-irrelevant distractors sampled from a restricted asset pool that excludes all object categories used as task-relevant objects anywhere in the benchmark. The OOD setting additionally randomizes scene backgrounds and illumination and introduces substantially denser clutter. OOD distractors are sampled from the full asset pool, excluding only objects already instantiated in the current scene and synonymous or near-duplicate assets, so objects that serve as task-relevant targets in other tasks can appear as distractors. Detailed evaluation configurations are provided in the appendix.

\noindent\textbf{Baselines and \mnm-VLA Suite.}
We evaluate six external baselines spanning three representative policy families. Large-scale pretrained VLA models include $\pi_{0.5}$~\citep{black2025pi05} and SaPaVe~\citep{liu2026sapave}. Memory-aware policies include HiF-VLA~\citep{lin2025hifvla}, MemoryVLA~\citep{shi2025memoryvla}, and MemER~\citep{sridhar2025memer}. We additionally evaluate Fast-WAM~\citep{yuan2026fastwam} as a representative world-action model.
We construct \mnm-VLA, a suite of 13 VLA configurations built upon the modular StarVLA~\citep{starvla2026} framework. An overview of the suite is shown in Fig.~\ref{fig:activearena_vla_suite}. \mnm-VLA is designed to isolate the effects of memory write policy, memory capacity, proprioceptive state input, auxiliary VLM supervision, high-level planning, and action decoding. \mnm-Fixed is a variant of \mnm-OFT that removes only the active-viewpoint action dimensions.
 Representative configurations are included in the main leaderboard, and the complete model specifications and evaluation details are provided in the appendix.
Table~\ref{tab:simulation_leaderboard} presents the simulation benchmark leaderboard. For each task category, ID and OOD success rates are reported side by side to characterize both in-distribution task learning and robustness to distribution shift. The results show that models with memory generally exhibit stronger generalization on active-perception tasks, and actively controllable viewpoint movement is essential.

\begin{table}[!htb]
\caption{
        Simulation benchmark leaderboard in task success rate (\%).
        Each category score is averaged over all tasks within that category,
        while Avg. is averaged over all 35 tasks.
        Best and second-best results are shown in bold and underlined, respectively.
    }
\label{tab:simulation_leaderboard}
    \centering
    \fontsize{9.1}{11}\selectfont
    % Allow space between numeric columns after parenthesizing citations.
    \setlength{\tabcolsep}{1.5pt}
    \renewcommand{\arraystretch}{1.12}

    \resizebox{\linewidth}{!}{%
    \begin{tabular}{
        @{}
        lc*{12}{c}
        @{}
    }
        \toprule
        \multirow{2}{*}{\textbf{Method}}
        & \multirow{2}{*}{\textbf{Memory}}
        & \multicolumn{2}{c}{\textbf{SS}}
        & \multicolumn{2}{c}{\textbf{SL}}
        & \multicolumn{2}{c}{\textbf{ML}}
        & \multicolumn{2}{c}{\textbf{MD}}
        & \multicolumn{2}{c}{\textbf{IA}}
        & \multicolumn{2}{c}{\textbf{Avg.}} \\
        \cmidrule(lr){3-4}
        \cmidrule(lr){5-6}
        \cmidrule(lr){7-8}
        \cmidrule(lr){9-10}
        \cmidrule(lr){11-12}
        \cmidrule(lr){13-14}
        &
        & ID & OOD
        & ID & OOD
        & ID & OOD
        & ID & OOD
        & ID & OOD
        & ID & OOD \\
        \midrule

        FAST-WAM~\citep{yuan2026fastwam}
        & No
        & 59.60 & 4.00
        & 40.38 & 1.00
        & 25.67 & 0.00
        & 12.00 & 3.33
        & 22.40 & 5.20
        & 35.60 & 2.06 \\

        $\pi_{0.5}$~\citep{black2025pi05}
        & No
        & 41.20 & 5.60
        & 22.38 & 0.63
        & 1.67 & 0.00
        & 2.67 & 5.33
        & 1.60 & 1.20
        & 16.86 & 1.71 \\

        SaPaVe~\citep{liu2026sapave}
        & No
        & 48.80 & 40.80
        & 19.88 & 5.88
        & 10.67 & 3.00
        & 25.33 & 12.67
        & 6.00 & 2.80
        & 20.91 & 10.51 \\

        \midrule

        HiF-VLA~\citep{lin2025hifvla}
        & Yes
        & 16.80 & 14.00
        & 6.88 & 1.13
        & 0.00 & 0.00
        & 30.67 & 18.67
        & 17.60 & 17.20
        & 10.69 & 6.57 \\

        MemER~\citep{sridhar2025memer}
        & Yes
        & \textbf{81.20} & 66.00
        & 33.88 & 18.63
        & 18.33 & 5.67
        & 40.00 & \textbf{42.00}
        & 11.60 & 6.40
        & 35.31 & 23.43 \\

        MemoryVLA~\citep{shi2025memoryvla}
        & Yes
        & 46.00 & 34.40
        & 20.88 & 6.50
        & 14.00 & 2.00
        & 28.00 & 27.33
        & 9.20 & 4.40
        & 22.23 & 11.20 \\

        \midrule

\mnm-Fixed
& Yes
& 1.20 & 0.40
& 0.00 & 0.00
& 0.00 & 0.00
& 1.33 & 0.67
& 0.40 & 0.40
& 0.34 & 0.17
\\

        \mnm-OFT
& Yes
& \underline{80.40} & \textbf{70.80}
& \textbf{74.50} & \textbf{54.38}
& \textbf{53.33} & \textbf{33.33}
& \textbf{44.00} & \underline{35.33}
& \textbf{31.60} & \textbf{27.20}
& \textbf{62.97} & \textbf{47.60} \\

        \mnm-Plan
        & Yes
        & 79.20 & \underline{69.60}
        & \underline{73.38} & \underline{52.63}
        & \underline{49.67} & \underline{30.00}
        & \underline{42.00} & 33.33
        & \underline{28.00} & \underline{24.80}
        & \underline{60.97} & \underline{45.54} \\

        \bottomrule
    \end{tabular}%
    }

\end{table}

\section{Analysis and Ablations}

\subsection{Failure-State Decomposition.}
We partition failed episodes into three mutually exclusive states. Discovery Miss (DM) indicates that the policy never observes all task-relevant objects; high DM rate therefore suggests that failures primarily arise from insufficient active perception. Context Omission (CO) indicates that all task-relevant objects were observed previously but are not jointly covered by the final observation; high CO rate suggests that failures mainly stem from the inability to actively retrieve relevant information from memory. Context Retained (CR) indicates that the final observation covers all task-relevant objects, yet the task still fails, suggesting errors in downstream decision-making or control. As shown in
Table~\ref{tab:failure_state_decomposition}, \mnm-OFT and \mnm-Plan
substantially reduce both DM and CO, demonstrating more reliable evidence
acquisition and availability at the final decision point.
Their remaining
failures are predominantly CR, indicating that the primary bottleneck
lies in downstream reasoning and execution rather than active perception. The results of \mnm-Fixed demonstrate that active viewpoint control is essential for completing the task.

\begin{table}[!htb]
\caption{Failure-state decomposition among unsuccessful episodes (\%):
discovery miss (DM), context omission (CO), and context retained (CR).
}
\label{tab:failure_state_decomposition}
    \centering
    \fontsize{9}{11}\selectfont
    \setlength{\tabcolsep}{1.0pt}
    \renewcommand{\arraystretch}{1.05}

    \begin{tabular*}{\columnwidth}{
        @{\extracolsep{\fill}}
        lc*{6}{c}
        @{}
    }
        \toprule
        \multirow{2}{*}{\textbf{Method}}
        & \multirow{2}{*}{\textbf{Mem.}}
        & \multicolumn{2}{c}{\textbf{DM}}
        & \multicolumn{2}{c}{\textbf{CO}}
        & \multicolumn{2}{c}{\textbf{CR}} \\
        \cmidrule(lr){3-4}
        \cmidrule(lr){5-6}
        \cmidrule(lr){7-8}

        & & ID & OOD
          & ID & OOD
          & ID & OOD \\
        \midrule

        FAST-WAM
        & No
        & 13.2 & 20.6
        & 34.6 & 47.2
        & 52.2 & 32.2 \\

        $\pi_{0.5}$
        & No
        & 20.9 & 25.2
        & 40.7 & 43.8
        & 38.4 & 31.0 \\

        SaPaVe
        & No
        & 12.6 & 13.9
        & 23.7 & 33.6
        & 63.7 & 52.5 \\

        \midrule

        HiF-VLA
        & Yes
        & 15.8 & 17.0
        & 21.4 & 20.9
        & 62.8 & 62.1 \\

        MemER
        & Yes
        & 9.1 & 13.5
        & 13.9 & 16.0
        & 77.0 & 70.5 \\

        MemoryVLA
        & Yes
        & 18.0 & 20.2
        & 21.3 & 17.3
        & 60.7 & 62.5 \\

        \midrule

        \mnm-Fixed
        & Yes
        & 96.3 & 97.3
        & 0.0 & 0.0
        & 3.7 & 2.7 \\

        \mnm-OFT
        & Yes
        & 2.3 & 3.2
        & 6.2 & 7.3
        & 91.5 & 89.5 \\

        \mnm-Plan
        & Yes
        & 3.0 & 4.7
        & 10.9 & 12.9
        & 86.1 & 82.4 \\

        \bottomrule
    \end{tabular*}
\end{table}

\subsection{Do the trends observed on \mnm-Bench transfer to real-world robotic manipulation?}
   We design eight real-world manipulation tasks and collect a total of 640 trajectories across them to train the evaluated policies. We then evaluate $\pi_{0.5}$, SaPaVe, MemER and \mnm-OFT
   on these tasks, conducting 20 evaluation trials per task for each policy. The results show that the memory-based model has a clear advantage on tasks requiring active perception, and the relative performance trends among the evaluated policies are consistent with those observed in \mnm-Sim. These findings indicate that {\mnm} captures challenges that are also relevant to physical robotic manipulation. Representative real-world rollouts are shown in Fig.~\ref{fig:real_rollout}. Detailed setup and task definitions are provided in the appendix.

% \begin{table}[t]
%   \centering
%   \scriptsize

%   \setlength{\tabcolsep}{3.2pt}
%   \renewcommand{\arraystretch}{1.12}
%   \begin{tabular}{p{0.35\columnwidth}cccc}
%     \toprule
%     Instruction &
%     $\pi_{0.5}$ &
%      SaPaVe &
%     MemER &
%     \shortstack{\mnm\\-OFT} \\
%     \midrule
%     Place pen in basket         & 45 & 55 & 65 & \textbf{75} \\
%     Place peach in basket       & 30 & 35 & 50 & \textbf{70} \\
%     Place screwdriver in basket & 25 & 15 & 30 & \textbf{55} \\
%     Place XL T-shirt in basket  &  0 &  5 & 10 & \textbf{40} \\
%     \midrule
%     Average                     & 25.0 & 27.5 & 38.8 & \textbf{60.0} \\
%     \bottomrule
%   \end{tabular}
%   % \vspace{-1mm}
%   \caption{
%     Real-world task success rates (\%). Instructions are shortened. All tasks require interacting with containers or occluding objects to locate the target object, followed by finding the basket and placing the object into it.
%   }
%   \label{tab:object_search_success}
%   \vspace{-5mm}
% \end{table}

\begin{reporttable}[0.85\linewidth]
\caption{
    Real-world task success rates (\%). Instructions are shortened.
    All tasks require interacting with containers or occluding objects
    to locate the target object, followed by finding the basket and
    placing the object into it. The average is computed over all eight tasks.
  }
\label{tab:object_search_success}
  \centering
  \small

  \setlength{\tabcolsep}{3.2pt}
  \renewcommand{\arraystretch}{1.12}
  \begin{tabular*}{\linewidth}{@{\extracolsep{\fill}}p{0.35\columnwidth}cccc}
    \toprule
    Instruction &
    $\pi_{0.5}$ &
    SaPaVe &
    MemER &
    \shortstack{\mnm-OFT} \\
    \midrule
    Place pen in basket
      & 45 & 55 & 65 & \textbf{75} \\
    Place peach in basket
      & 30 & 35 & 50 & \textbf{70} \\
    Place screwdriver in basket
      & 25 & 15 & 30 & \textbf{55} \\
    Place XL T-shirt in basket
      &  0 &  5 & 10 & \textbf{40} \\

   \multicolumn{5}{c}{\ldots}  \\

    \midrule
    Average
      & 23.8 & 24.4 & 34.4 & \textbf{55.6} \\
    \bottomrule
  \end{tabular*}

\end{reporttable}

\begin{figure}[!tp]
    \centering
    \includegraphics[width=\columnwidth]{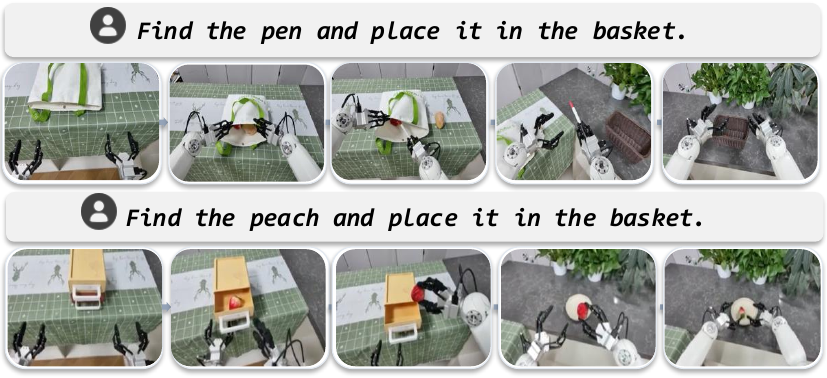}
    \caption{Two representative rollouts from two of the eight real-world manipulation tasks.}
    \label{fig:real_rollout}
    
\end{figure}

\subsection{Effect of Memory Write Policy and Capacity.}
Perceptual memory is jointly governed by its write policy and capacity. We evaluate three memory write policies: \emph{subtask}, which writes a frame when the predicted subtask changes; \emph{motion}, which writes a frame upon gripper transitions or thresholded end-effector, head, or torso motion; and \emph{chunk}, which writes one frame after each action chunk. Combining each policy with memory capacities of 6 and 12 frames yields six \mnm-OFT variants. We also include a memory-free OFT~\citep{kim2025oft} baseline. All variants share the same task instruction, 18-D proprioceptive state, OFT action head, and subtask-supervised VLM objective. Results are reported in Table~\ref{tab:memory_write}.

\begin{table}[!htb]
\caption{
        Effect of memory write policy and capacity on task success rate (\%).
        Category scores are averaged over their corresponding tasks, while
        Avg.\ is the task-macro average over all 35 tasks
        (5 SS, 16 SL, 6 ML, 3 MD, and 5 IA).
        Best and second-best results are shown in bold and underlined,
        respectively.
    }
\label{tab:memory_write}
    \centering
    \footnotesize
    \setlength{\tabcolsep}{2.2pt}
    \renewcommand{\arraystretch}{1.05}

    \begin{tabular*}{\textwidth}{
        @{\extracolsep{\fill}}
        ll*{12}{c}
        @{}
    }
        \toprule
        \multirow{2}{*}{\shortstack[l]{\textbf{Memory Write}\\\textbf{Policy}}}
        & \multirow{2}{*}{\textbf{Capacity}}
        & \multicolumn{2}{c}{\textbf{SS}}
        & \multicolumn{2}{c}{\textbf{SL}}
        & \multicolumn{2}{c}{\textbf{ML}}
        & \multicolumn{2}{c}{\textbf{MD}}
        & \multicolumn{2}{c}{\textbf{IA}}
        & \multicolumn{2}{c}{\textbf{Avg.}} \\
        \cmidrule(lr){3-4}
        \cmidrule(lr){5-6}
        \cmidrule(lr){7-8}
        \cmidrule(lr){9-10}
        \cmidrule(lr){11-12}
        \cmidrule(lr){13-14}

        & & ID & OOD
          & ID & OOD
          & ID & OOD
          & ID & OOD
          & ID & OOD
          & ID & OOD \\
        \midrule

        None
        & 0
        & 58.80 & 15.20
        & 27.63 & 2.13
        & 13.67 & 0.00
        & 36.67 & 24.00
        & 6.00 & 6.40
        & 27.37 & 6.12 \\

        \midrule

        \multirow{2}{*}{Subtask}
        & 6
        & \underline{66.40} & 60.80
        & 29.50 & 19.13
        & 0.00 & 0.00
        & 16.00 & 13.33
        & 0.80 & 0.80
        & 24.46 & 18.69 \\

        & 12
        & 62.80 & 54.40
        & 37.25 & 15.00
        & 0.67 & 0.33
        & 21.33 & 15.33
        & 1.20 & 0.00
        & 28.11 & 16.00 \\

        \midrule

        \multirow{2}{*}{Motion}
        & 6
        & 62.40 & 54.80
        & 45.63 & 28.50
        & 20.33 & 10.67
        & 36.00 & 32.67
        & 8.40 & 6.40
        & 37.54 & 26.40 \\

        & 12
        & 58.00 & 47.60
        & 36.75 & 20.75
        & 24.00 & 11.67
        & 22.00 & 30.67
        & 12.00 & 12.40
        & 32.80 & 22.69 \\

        \midrule

\multirow{2}{*}{Chunk}
& 6
& \textbf{80.40} & \underline{63.60}
& \underline{65.25} & \underline{40.88}
& \underline{49.00} & \underline{21.67}
& \textbf{50.67} & \textbf{46.00}
& \textbf{32.40} & \underline{23.20}
& \underline{58.69} & \underline{38.74} \\

        & 12
        & \textbf{80.40} & \textbf{70.80}
        & \textbf{74.50} & \textbf{54.38}
        & \textbf{53.33} & \textbf{33.33}
        & \underline{44.00} & \underline{35.33}
        & \underline{31.60} & \textbf{27.20}
        & \textbf{62.97} & \textbf{47.60} \\

        \bottomrule
    \end{tabular*}

\end{table}

\begin{figure}[!tp]
    \centering
    \includegraphics[width=0.82\linewidth]{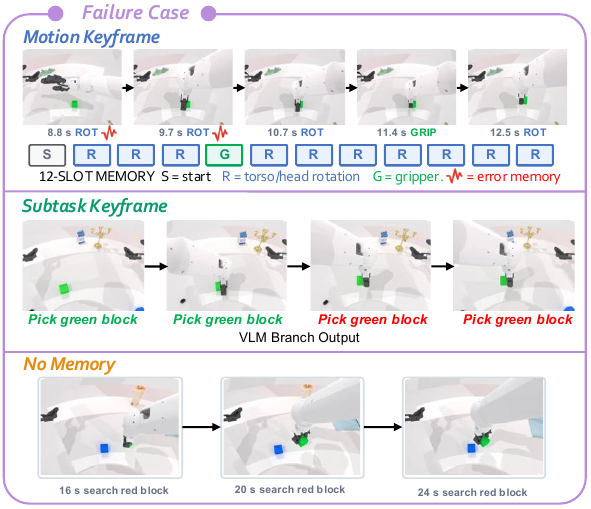}
    \caption{Failure cases of three memory writing policies.}
    \label{fig:failure_cases}
    
\end{figure}

As shown in Fig.~\ref{fig:failure_cases}, the memory-free policy cannot track explored regions or plan subsequent searches, highlighting the importance of memory for closed-loop active perception. This finding is consistent with the results in Table~\ref{tab:simulation_leaderboard}.
The two event-triggered write policies show limited robustness. The subtask-based policy degrades substantially from SS to IA and ML. Its sparse, prediction-dependent triggers can mistime memory updates and omit informative views. The motion-based policy is sensitive to control jitter, which introduces redundant or irrelevant frames. Increasing memory capacity retains more of this noise and reduces performance in most settings. Both policies degrade further under OOD evaluation, indicating their sensitivity to prediction errors and control noise.
The chunk-based write policy achieves the best performance across all task--evaluation pairs. Its regular update schedule avoids semantic predictions and motion thresholds, providing stable temporal coverage and stronger OOD robustness. Increasing its capacity from 6 to 12 frames improves six of ten settings, with the largest gains on OOD SL and ML. A reliable write policy can thus exploit additional capacity to improve generalization on complex tasks.
Overall, the memory write policy is more critical than capacity alone. Additional capacity yields consistent gains only when task-relevant observations are written reliably.

\subsection{Planner-Mediated Memory Management.}
Recently, many models adopt hierarchical frameworks in which a high-level planner manages memory inputs and decomposes tasks~\citep{shi2025hirobot,sridhar2025memer,chen2026rmbench}. To evaluate the effectiveness of planner-based architectures, we introduce \mnm-Plan.  The planner is fine-tuned from Qwen3-VL-2B~\citep{bai2025qwen3vl}  and takes as input the global task instruction, the current observation, and a rolling history of up to 12 frames sampled at action-chunk boundaries. It predicts the current subtask instruction and selects a sparse subset of candidate frames for long-term memory. The downstream VLA then predicts the next action chunk conditioned on the current observation, selected memory frames, predicted subtask instruction, and proprioceptive state. Its architecture is illustrated in Fig.~\ref{fig:planner}. Architectural and training details are provided in the appendix.
%
% The results (Table~\ref{tab:planner_memory}) show that delegating subtask inference and memory writing to the planner is more stable than jointly coupling subtask-transition prediction and memory management with the low-level VLA policy, while achieving performance close to our best-performing variant using only sparse memory. \mnm-Plan retains near-best performance with sparse memory, while reducing peak GPU memory usage during VLA training by 46\%.

The results in Table~\ref{tab:planner_memory} show that planner-mediated subtask inference and memory writing are more stable than jointly coupling these functions with low-level policy, while retaining near-best performance with sparse memory and reducing peak GPU memory usage during VLA training by 46\%.

\begin{figure}[!tp]
    \centering
    \includegraphics[width=\columnwidth]{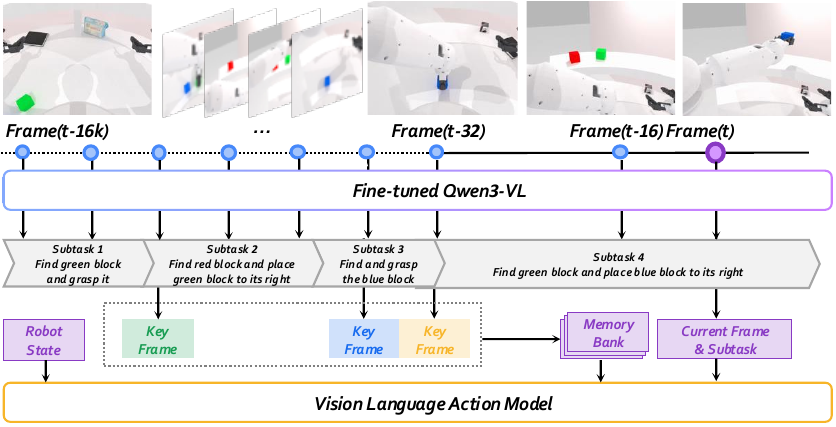}
    \caption{\mnm-Plan. The Planner identifies historical frames that span subtask transitions and adds them to the memory bank, while inferring the current subtask.
}
    \label{fig:planner}
    
\end{figure}

\begin{reporttable}[0.85\linewidth]
\caption{
        Comparison of planner-mediated memory management and memory write
        policies on task success rate (\%).
    }
\label{tab:planner_memory}
    \centering
    \small
    \setlength{\tabcolsep}{5.5pt}
    \renewcommand{\arraystretch}{1.12}

    \begin{tabular*}{\linewidth}{@{\extracolsep{\fill}}lccc}
        \toprule
        \textbf{Method}
        & \textbf{Memory Management}
        & \textbf{ID}
        & \textbf{OOD} \\
        \midrule

       \mnm-OFT
        & Chunk Writing
        & \textbf{62.97}
        & \textbf{47.60} \\

       \mnm-OFT
        & Subtask Writing
        & 28.11
        & 16.00 \\

       \mnm-Plan
        & Planner Selection
        & \underline{60.97}

        & \underline{45.54} \\

        \bottomrule
    \end{tabular*}

\end{reporttable}

\Needspace{8\baselineskip}
\subsection{Effect of VLM Branch Supervision.}
To examine whether stronger task understanding in the VLM branch improves policy generalization, we compare three \mnm-VLA variants: no auxiliary VLM supervision, task-level instruction supervision, and subtask-level supervision. All other components are held fixed, including the OFT architecture, 12-frame Chunk memory, and explicit proprioceptive state conditioning. The results in Table~\ref{tab:vlm_supervision} show that fine-grained subtask supervision improves generalization under changes in scene appearance and distractor composition.

\begin{reporttable}[0.85\linewidth]
\caption{
        Comparison of different auxiliary VLM supervision on task success rate (\%). All variants' VLM branch uses next-token-prediction loss while keeping the same action loss.
    }
\label{tab:vlm_supervision}
    \centering
    \small
    \setlength{\tabcolsep}{6pt}
    \renewcommand{\arraystretch}{1.12}

    \begin{tabular*}{\linewidth}{@{\extracolsep{\fill}}llcc}
        \toprule
        \textbf{Method}
        & \textbf{VLM Supervision}
        & \textbf{ID}
        & \textbf{OOD} \\
        \midrule

        \multirow{3}{*}{\mnm-OFT}
        & None
        & 58.57
        & \underline{41.60} \\

        & Task-level
        & \underline{60.06}
        & 39.54\\

        & Subtask-level
        & \textbf{62.97}
        & \textbf{47.60} \\

        \bottomrule
    \end{tabular*}

\end{reporttable}

\Needspace{8\baselineskip}
\subsection{Effect of Proprioceptive State Conditioning.}
\begin{figure}[!tp]
    \centering
    \includegraphics[width=\columnwidth]{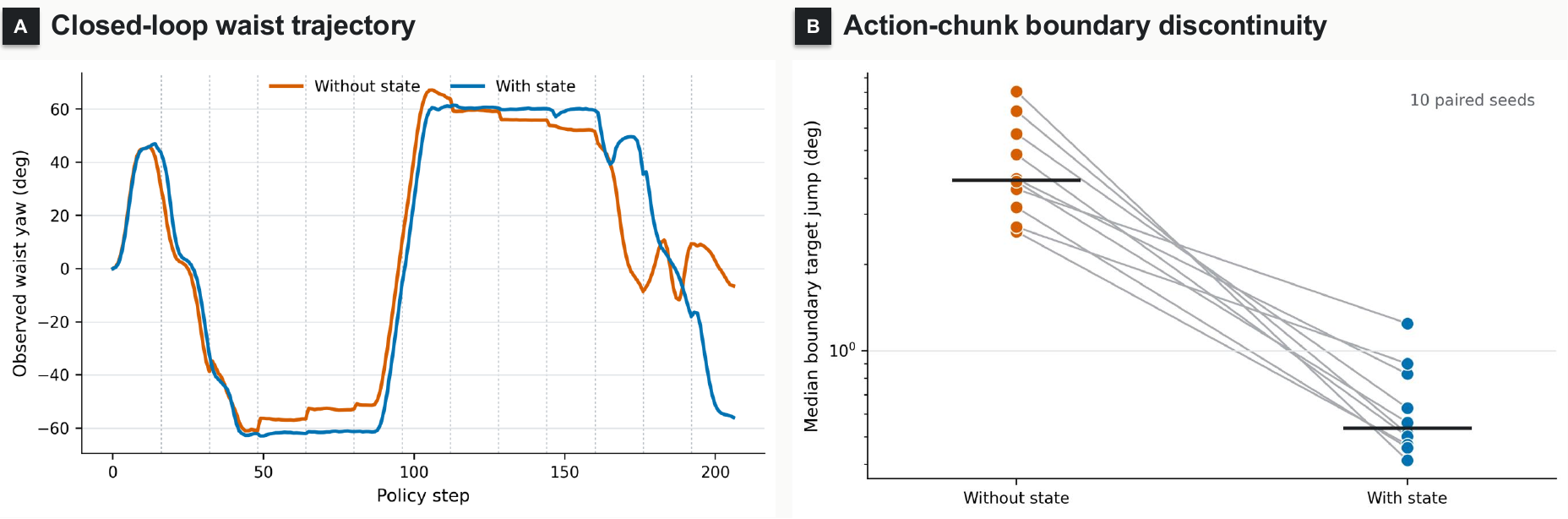}
    \caption{
    Effect of proprioceptive state conditioning on action continuity.
    Left: torso target trajectory, with dashed lines indicating 16-step action-chunk boundaries.
    Right: episode-level median target discontinuity at chunk boundaries across 10 paired seeds; gray lines connect identical seeds.
    }
    \vs
    \label{fig:state_ablation}
\end{figure}

For each visual frame, we encode the corresponding 18-D proprioceptive state with an MLP and fuse the resulting state embedding with the visual and language tokens~\citep{yuan2026qwen}. We ablate proprioceptive state conditioning while holding the remaining \mnm-OFT configuration fixed. Table~\ref{tab:robot_state} reports the task success rates, and Fig.~\ref{fig:state_ablation} analyzes its effect on closed-loop action continuity.
\begin{reporttable}[0.85\linewidth]
\caption{
        Comparison of explicit proprioceptive state conditioning on task success rate (\%).
    }
\label{tab:robot_state}
    \centering
    \small
    \setlength{\tabcolsep}{7pt}
    \renewcommand{\arraystretch}{1.12}

    \begin{tabular*}{\linewidth}{@{\extracolsep{\fill}}llcc}
        \toprule
        \textbf{Method}
        & \textbf{State Input}
        & \textbf{ID}
        & \textbf{OOD} \\
        \midrule

        \multirow{2}{*}{\mnm-OFT}
        & No
        & 50.86
        & 16.86 \\

        & Yes
        & \textbf{62.97}
        & \textbf{47.60} \\

        \bottomrule
    \end{tabular*}

\end{reporttable}

To examine this effect at the action level, we evaluate both variants on the \emph{place-object-stand-rotate-view} task using the same 10 random seeds. Fig.~\ref{fig:state_ablation} shows that proprioceptive state conditioning substantially reduces discontinuities at action-chunk boundaries, lowering the mean episode-level median target jump from $4.55^\circ$ to $0.65^\circ$. Explicit state information anchors each predicted chunk to the current robot configuration, reducing the ambiguity of inferring posture from visual observations and producing smoother closed-loop control.

% proprioceptive state conditioning increases the average success rate from $36.07\%$ to $50.55\%$. The gain is concentrated under distribution shift: the average OOD success rate rises from $19.52\%$ to $44.21\%$, with improvements across all five task categories. The largest gains occur on SS, SL, and ML, reaching $40.40$, $40.38$, and $31.33$ percentage points, respectively. ID performance is mixed, with gains on SS, SL, and ML but declines on MD and IA. These results indicate that explicit state information primarily strengthens OOD robustness rather than uniformly improving in-distribution performance.

\Needspace{8\baselineskip}
\subsection{Effect of Action Head.}
To isolate the effect of action generation, we compare three representative action heads while holding the VLM backbone, memory configuration, training data, and input modalities fixed. The controls for both arms, both grippers, the torso, and the head are jointly represented as an 18-D action vector. 
OFT~\citep{kim2025oft} uses an MLP to regress continuous action chunks in parallel from action query token representations. FAST~\citep{pertsch2025fast} discretizes action chunks and predicts the resulting action tokens autoregressively. GR00T~\citep{nvidia2025gr00tn1} uses a DiT-based flow-matching expert to generate continuous action chunks. Table~\ref{tab:action_head} reports their ID and OOD performance. GR00T's failures mainly stem from action noise, which typically requires large-scale pretraining to mitigate, whereas FAST's failures are primarily caused by decoding instability.

\begin{reporttable}[0.85\linewidth]
\caption{
        Comparison of different action-head architectures on task success rate (\%). 
    }
\label{tab:action_head}
    \centering
    \small
    \setlength{\tabcolsep}{8pt}
    \renewcommand{\arraystretch}{1.12}

    \begin{tabular*}{\linewidth}{@{\extracolsep{\fill}}lcc}
        \toprule
        \textbf{Method}
        & \textbf{ID}
        & \textbf{OOD} \\
        \midrule

\mnm-OFT
& \textbf{62.97}
& \textbf{47.60} \\

\mnm-GR00T
& \underline{40.00}
& \underline{13.43} \\

\mnm-FAST
& 22.34
& 12.00 \\

        \bottomrule
    \end{tabular*}

\end{reporttable}

\section{Conclusion}

In this work, we present {\mnm}, a comprehensive simulator--benchmark--baseline suite comprising \mnm-Sim, \mnm-Bench, and \mnm-VLA to study active perception and manipulation.
Our systematic analysis reveals that principled memory and model design---including memory capacity, maintenance strategies, multimodal integration, supervision, and hierarchical architectures---is fundamental to acquiring and retaining task-relevant information over long horizons.
Despite these advances, a substantial ID--OOD gap persists, highlighting the inherent challenges of active perception and manipulation and motivating future research.
Future work may explore more effective multimodal memory mechanisms and more scalable system architectures, as well as pre-training paradigms that endow embodied agents with active perception and manipulation capabilities from large-scale egocentric data.
% Supplementary material is part of the same report and uses alphabetic
% sectioning after the main text.
\clearpage
\bibliographystyle{iclr2027_conference}
\bibliography{references}
\clearpage
\appendix
\section*{Appendix}
\addcontentsline{toc}{section}{Appendix}
\section{Related Work}

\paragraph{Robot Manipulation, Active Perception, and Memory Benchmarks.}
Robot manipulation benchmarks have evolved from standardized multi-task control toward language-conditioned, long-horizon, household-scale, and distribution-aware evaluation.
Meta-World, RLBench, and ManiSkill established general-purpose testbeds for manipulation learning and transfer~\citep{yu2020meta,james2020rlbench,gu2023maniskill2}; CALVIN, VIMA, and LIBERO extended this scope to language grounding and compositional task learning~\citep{mees2022calvin,jiang2023vimageneralrobotmanipulation,liu2023libero}; and recent platforms such as BEHAVIOR-1K, RoboCasa365, RoboTwin~2.0, RoboDojo, and LIBERO-Plus further emphasize household diversity, bimanual control, sim-to-real deployment, and robustness to distribution shifts~\citep{li2024behavior,nasiriany2026robocasa365,chen2025robotwin2,chen2026robodojo,fei2025liberoplus}.
Despite this progress, most manipulation benchmarks assume that task-relevant evidence is already available in the observation stream and therefore evaluate how a policy acts on perceived information rather than how it acquires missing evidence.
Active-perception benchmarks instead treat sensing as part of decision making: EmbodiedQA, OpenEQA, EmbodiedEval, CHAIN, and ESI-Bench require agents to explore or interact with 3D environments to resolve uncertainty~\citep{das2018embodied,OpenEQA2023,cheng2025embodiedeval,wu2026chain,hong2026esibench}.
However, these benchmarks primarily operate through high-level semantic actions and largely abstract away continuous robot control.
In parallel, memory-centric manipulation suites evaluate delayed evidence, state aliasing, and history-dependent decisions~\citep{fang2025sam2act,cherepanov2025mikasa,lei2026robomemarena,dai2026robomme,chen2026rmbench,shah2026prism}
%, while ActiveManip-Bench begins to couple policy-controlled sensing with dual-arm manipulation~\citep{liu2026sapave}
. These directions remain only partially unified: memory benchmarks emphasize retaining acquired evidence, whereas active perception additionally requires deciding what evidence to acquire and how to obtain it through closed-loop action.
{\mnm} integrates incomplete initial evidence, active viewpoint control, physical information acquisition, spatiotemporal memory, and low-level dual-arm manipulation within a standardized ID/OOD evaluation framework.

\paragraph{Vision--Language--Action Models and Memory.}
Vision--language--action (VLA) models have recently enabled generalist robot policies by aligning visual, linguistic, and action representations.
Representative systems, including RT-2, Octo, OpenVLA, $\pi_0$, $\pi_{0.5}$, and GR00T, explore large-scale pretraining, heterogeneous robot data, and expressive action generation~\citep{zitkovich2023rt,team2024octo,kim2024openvla,black2025pi0,black2025pi05,nvidia2025gr00tn1}. Related VLM-based approaches address spatial referring, spatial tracing, and geometric reasoning for robotic manipulation~\citep{zhou2026roborefer,zhou2026towards,han2025tiger}.
Beyond action generation, recent efforts increasingly investigate how VLAs utilize temporal information.
Different approaches construct memory through recent visual traces, explicit keyframe banks, semantic retrieval, or recurrent latent states, including TraceVLA, HiF-VLA, MemoryVLA, MemER, and ReMemVLA~\citep{zheng2024tracevla,lin2025hifvla,shi2025memoryvla,sridhar2025memer,li2026remem}. 
These methods demonstrate that retaining historical evidence improves long-horizon manipulation, but they mainly focus on memory utilization after observations are acquired.
A complementary direction introduces active sensing policies that explicitly control viewpoints or exploration behaviors, such as SaPaVe and CoMe-VLA~\citep{liu2026sapave,li2026act}. Visual feedback also supports execution progress estimation~\citep{tan2026robobrain} and constraint-aware failure monitoring~\citep{zhou2024code} in robotic manipulation.
Nevertheless, existing VLA evaluations lack a unified framework for studying how a policy decides what information to acquire, how to encode acquired evidence, and how memory interacts with low-level manipulation.
{\mnm} provides such a testbed by combining active information acquisition, diverse memory annotations, and continuous robot control within a unified benchmark.

\section{Details of \mnm-Sim}

This section summarizes the engineering modifications used to transform RoboTwin~2.0 into \mnm-Sim. We focus on extensions to the robot embodiment, workspace, task assets, action interface, and annotation pipeline; standard RoboTwin data-generation, domain-randomization, and manipulation APIs are omitted.

\subsection{Astribot Embodiment Adaptation}

\begin{figure}[!tp]
    \centering
    \includegraphics[width=\linewidth]{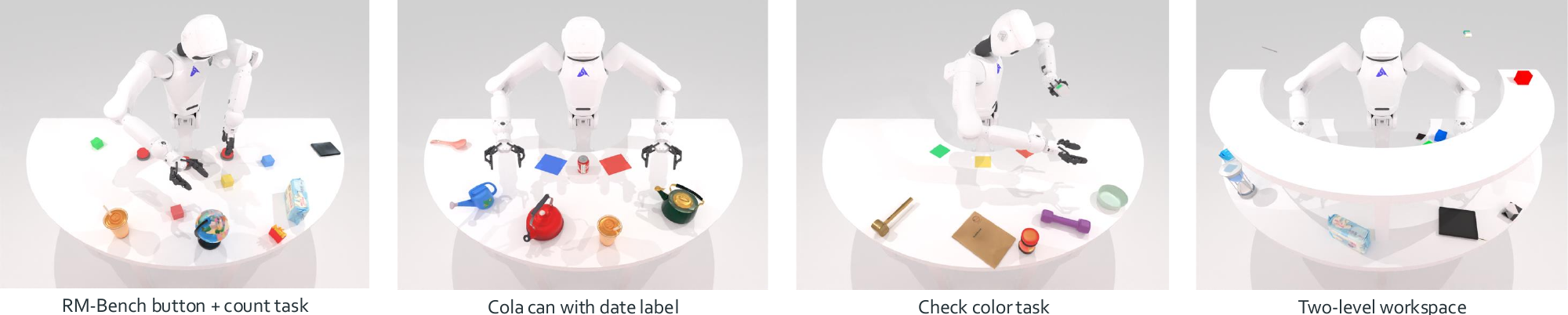}
    \caption{Representative simulator frames.  The panels show the RM-Bench
  button integrated into a counting task, a block with hidden color patch, a
  can with a runtime-generated date label, and the two-level fan workspace.
}
    \label{fig:scene_examples}
\end{figure}
We add an Astribot S1 whole-body URDF, a manually annotated CuRobo self-collision model, and independent CuRobo planning models for the two arms. The robot is loaded as a single fixed-base articulation at
\[
[x,y,z,q_w,q_x,q_y,q_z]
=[0,-0.45,-0.15,0.707,0,0,0.707].
\]
This transform aligns the native robot frame with RoboTwin's world frame and places the base at the center of the fan-shaped workspace. The model comprises two seven-DoF arms, a single-DoF head that rotates vertically in pitch, an actuated torso-yaw joint, and two parallel grippers. We replace the original multi-link gripper actuation with a single-command mimic-joint model: one active joint drives five mimic joints, enabling scalar open/close control.
 
\begin{table}[!htb]
\caption{Astribot control parameters introduced in \mnm-Sim. Angles are in radians. Gripper values follow RoboTwin's normalized convention, where 0 and 1 denote closed and open, respectively.}
\label{tab:astribot-parameters}
  \centering
  \small

  \begin{tabularx}{\linewidth}{@{}l l X@{}}
    \toprule
    Component & Parameter & Value / implementation \\
    \midrule
    Left arm & Home state & $[-0.4,-0.7,-1.0,1.4,-0.9,0,0]$ \\
    Right arm & Home state & $[0.4,-0.7,1.0,1.4,0.9,0,0]$ \\
    Head & Joint limits & pitch $[-1.22,1.22]$\\
    Head & Home; velocity / acceleration & $1.0$; $10$ rad/s and $25$ rad/s$^2$ \\
    Torso & Joint and limits & \texttt{astribot\_torso\_}\allowbreak\texttt{joint\_4}, $[-1.2,1.2]$ \\
    Torso & Home; velocity / acceleration & $0$; $2.5$ rad/s and $10$ rad/s$^2$ \\
    Arm drives & Stiffness / damping & $2000/200$ \\
    Torso drive & Stiffness / damping & $3000/800$ \\
    Gripper drive & Stiffness / damping / force & $4000/200/120$ \\
    Gripper mapping & Normalized command to active joint & \texttt{gripper\_scale}$=[-1,0]$; bias $0.21$ m \\
    \bottomrule
  \end{tabularx}
\end{table}

The motion planner uses \texttt{astribot\_arm\_left\_link\_7} and \texttt{astribot\_arm\_right\_link\_7} as the left and right move groups. We calibrate RoboTwin's tool-center-point convention against these end links using
\begin{equation}
  R_{\Delta}=
  \begin{bmatrix}
    0&-1&0\\
    -1&0&0\\
    0&0&-1
  \end{bmatrix},\qquad
  R_{G}=
  \begin{bmatrix}
    0&0&1\\
    0&1&0\\
    -1&0&0
  \end{bmatrix}.
\end{equation}
Before planning, the TCP is shifted by $0.12-0.21=-0.09$ m along the local tool $x$ axis. Calibration is estimated from paired TCP and end-link poses and verified by world-frame forward kinematics. We additionally implement joint-specific drive overrides, joint-limit clipping, synchronized initialization of articulation states and drive targets, and collision filtering between the head and both arms.

\subsection{Moving Cameras and Calibration}
\label{app:cameras}

We add one massless head-camera link and two massless wrist-camera links to the URDF. The head camera is attached to \texttt{astribot\_head\_}\allowbreak\texttt{link\_2} with translation $[0.066,-0.210,0]$ m and roll--pitch--yaw $[\pi/2,0,0]$. Each wrist camera is attached to its gripper base with translation $[0,-0.05661,0.01979]$ m and roll--pitch--yaw $[0,-1.37081,1.5708]$. Although these parent-to-camera transforms are fixed, the world extrinsics depend on the robot configuration:
\begin{equation}
  {}^{W}T_{C}(t) = {}^{W}T_{P}(q_t)\,{}^{P}T_{C}.
  \label{eq:moving-camera}
\end{equation}
Camera poses are therefore updated through forward kinematics and recorded per frame. The head camera moves only with the single pitch joint, while horizontal viewpoint changes are produced by torso yaw.

The policy head camera uses a new preset: $512\times384$ pixels, a $60^\circ$ vertical field of view, and 0.1--100 m near/far planes. Under an ideal pinhole model, this gives $f_x=f_y=332.55$ px and $(c_x,c_y)=(256,192)$ px. Optional wrist cameras use $320\times240$ pixels and a $37^\circ$ vertical field of view ($f_x=f_y\simeq358.64$ px). The benchmark disables wrist streams and exposes only the moving head camera to the policy. 

\subsection{Fan-Shaped and Two-Level Workspaces}
\label{app:workspace}

\begin{figure}[!tp]
    \centering
    \includegraphics[width=\linewidth]{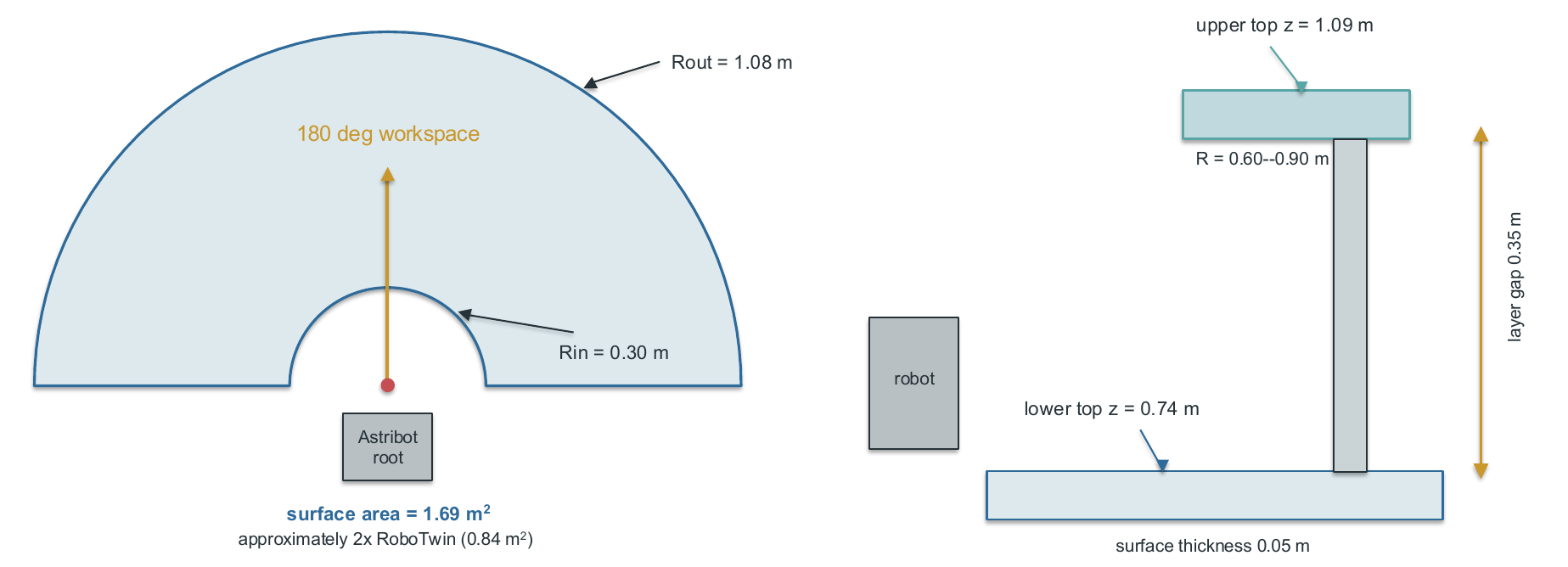}
    \caption{Fan-Shaped and Two-Level Workspaces.}
    \label{fig:workspaces}
\end{figure}

A rectangular table keeps most objects within a single frontal view and is therefore poorly suited to active visual search. We instead construct an annular fan-shaped table centered at the robot base. Its visual and PhysX collision surfaces are generated from the same thin box patches, with 14 radial segments, at least 24 angular segments, and a target density of 18 segments/m. The benchmark uses inner and outer radii of 0.30 and 1.08 m, a $180^\circ$ span centered at $90^\circ$ in world coordinates, a $20^\circ$ placement margin at each boundary, a table thickness of 0.05 m, and a lower tabletop height of 0.74 m.

The two-level variant retains the lower fan and adds an upper annular band with inner/outer radii of 0.60/0.90 m. A 0.35 m inter-level gap places the upper surface at 1.09 m. Its angular interval may be shifted by $20^\circ$ to induce asymmetric occlusion, and a support column is placed at the center of the upper span. Clutter placement is correspondingly changed from rectangular $x$--$y$ sampling to annular sampling.

\subsection{Additional Assets and Tasks}
\label{app:assets}

\paragraph{Button asset.}
The three count-and-press tasks use the articulated \texttt{005\_button} asset from RM-Bench. A button is considered pressed when its joint position falls below $-0.005$ m and reset when it rises above $-0.001$ m.

\paragraph{Task-specific assets.}
For check-color tasks, we add a thin colored patch to the rear face of an otherwise gray dynamic block. The block half-size is 0.024 m; the patch half-size is $(0.0015,0.012,0.012)$ m and protrudes by only 0.00005 m.

For can-inspection tasks, we retain the collision and primary visual geometry of RoboTwin's \texttt{071\_can/}\allowbreak\texttt{base3} but generate the information-bearing surface ourselves. The date task creates and caches an OBJ/MTL/PNG curved label at runtime: a $45^\circ$ cylindrical arc of radius 0.029 m centered on the rear face. Production dates range from 2023-01-01 to 2026-12-31; valid and expired samples are balanced under a 365-day shelf life. The color-check can task uses the same mechanism to generate a colored rear label.

\subsection{New Action Primitives}
\label{app:action-primitives}

We retain RoboTwin's Cartesian arm-motion and gripper primitives and add the controls required for embodiment-level motion and active perception:
\begin{itemize}
  \item \texttt{move\_joint(target\_joint\_pos)} commands one arm in joint space. Missing target entries are filled from the current state, excess entries are truncated, and all joints are clipped to URDF limits. TOPP time parameterization is used when available; otherwise, the fallback is 1 rad/s linear interpolation over at least 20 simulation steps.
  \item \texttt{move\_head(delta)} and \texttt{move\_head\_to(target)} provide relative and absolute control of the single head-pitch joint. Likewise, \texttt{move\_torso(delta)} and \texttt{move\_torso\_to(target)} control torso yaw. Both use joint-limit-clipped triangular or trapezoidal velocity profiles at the simulator timestep.
  \item \texttt{look\_at\_world\_point\_with\_}\allowbreak\texttt{head} and \texttt{look\_at\_object} convert a world point, object center, contact point, or functional point into a viewing target. Horizontal orientation is assigned to \texttt{astribot\_torso\_}\allowbreak\texttt{joint\_4}, while head pitch controls elevation.
  \item \texttt{face\_world\_point\_with\_}\allowbreak\texttt{torso}, \texttt{face\_object\_with\_}\allowbreak\texttt{torso}, and \\ \texttt{search\_and\_focus\_}\allowbreak\texttt{rotate\_subtask} implement discrete sector search, target centering, and visibility/memory updates.
\end{itemize}
Head or torso commands cannot be combined with arm/gripper commands within the same primitive step. The action vector is 18-dimensional, comprising 14 arm joints, two grippers, torso yaw, and head pitch.

\subsection{Per-Frame Data and Online Annotations}
\label{app:frame-schema}

At each simulator step, the benchmark records core observations, actions, and task metadata in HDF5. Table~\ref{tab:trajectory-schema} summarizes the training-facing schema, where $N$ is the number of registered task objects.

\begin{table}[!htb]
\caption{Per-frame trajectory fields recorded by the benchmark configuration.}
\label{tab:trajectory-schema}
  \centering
  \scriptsize

  \begin{tabularx}{\linewidth}{@{}l l l X@{}}
    \toprule
    Group & Field & Shape / type & Meaning \\
    \midrule
    \texttt{observation/camera\_head} & \texttt{rgb} & encoded RGB & moving head-camera image \\
     & \texttt{intrinsic\_cv} & $3\times3$, float & per-frame camera intrinsics \\
     & \texttt{extrinsic\_cv} & $4\times4$, float & world-to-camera transform, CV convention \\
     & \texttt{cam2world\_gl} & $4\times4$, float & camera-to-world transform, OpenGL convention \\
    \texttt{endpose} & \texttt{left/right\_endpose} & $7$, float & TCP position and quaternion \\
     & \texttt{left/right\_gripper} & scalar, float & normalized gripper state \\
    \texttt{joint\_action} & \texttt{left/right\_arm} & $7$, float & arm joint state \\
     & \texttt{left/right\_gripper} & scalar, float & normalized gripper state \\
     & \texttt{head/torso} & $1$ / $1$, float & head pitch and torso-yaw state \\
     & \texttt{vector} & $18$, float & concatenated simulator state \\
    Top level & \texttt{subtask} & int32 & current semantic subtask ID \\
     & \texttt{stage} & int8 & 1: search; 2: anchor; 3: action \\
     & \texttt{subtask\_instruction\_idx} & int32 & instruction lookup ID \\
     & \texttt{focus\_object\_idx} & int16 & focused object; $-1$ if absent \\
     & \texttt{focus\_object\_visible} & int8 & whether the focused object is currently visible \\
     & \texttt{info\_complete} & int8 & whether sufficient information has been acquired \\
     & \texttt{camera\_mode} & int8 & active-camera state-machine mode \\
     & \texttt{camera\_target\_theta} & float32 & planned viewing angle; NaN if inactive \\
     & \texttt{waist\_heading\_deg} & float32 & current horizontal viewing direction \\
     & \texttt{visible\_object\_mask} & $N$, int8 & objects visible in the current frame \\
     & \texttt{discovered\_object\_mask} & $N$, int8 & objects observed at least once \\
     & \texttt{search\_target\_mask} & $N$, int8 & objects required for information acquisition \\
     & \texttt{action\_target\_mask} & $N$, int8 & objects involved in the physical action \\
     & \texttt{carried\_object\_mask} & $N$, int8 & objects currently carried \\
     & \texttt{target\_uv\_norm} & $2$, float32 & normalized target coordinate; $(-1,-1)$ if absent \\
    Post-process & \texttt{subtask\_keyframe} & int8 & semantic-boundary indicator \\
     & \texttt{motion\_keyframe} & int8 & causal motion-boundary indicator \\
    \bottomrule
  \end{tabularx}
\end{table}

Each episode also includes a human-readable JSON sidecar containing the task instruction, object key/index/name mappings, subtask definitions, transition log, and final discovery state. All annotations are generated online from simulator and task state rather than labeled after collection.

\section{\mnm-Bench Design and Data Protocol}
\label{app:bench-design}

This section describes the design principles, evaluation domains, data protocol, and history-keyframe definitions of our information-gathering manipulation benchmark. Unlike conventional tasks that can be solved from the current observation alone, these tasks require the robot to actively acquire missing evidence, retain it after it leaves the field of view, and use it to guide subsequent decisions.

\subsection{Task Design Principles and Object Distribution}
\label{app:task-principles}

All tasks satisfy four design principles.
\textbf{(1)} \textit{Initial Information Insufficiency.}
The correct manipulation cannot be determined from the initial observation alone.
\textbf{(2)} \textit{Information Accessibility.}
The missing task-relevant information can be acquired through spatial search or physical interaction.
\textbf{(3)} \textit{Historical Dependence.}
Evidence acquired from previous observations must remain relevant even after the corresponding objects or regions leave the current field of view.
\textbf{(4)} \textit{Closed-Loop Decision Making.}
Newly acquired evidence must influence subsequent localization or manipulation decisions.
After each action, the robot must reassess whether the accumulated evidence is sufficient and continue acquiring information when necessary.

These principles jointly constrain scene generation and object placement. Figure~\ref{fig:workspaces} illustrates the shared distribution template. Single-level tasks use a $180^\circ$ annular sector centered at the robot, with inner and outer radii of $0.3\,\mathrm{m}$ and $1.08\,\mathrm{m}$, respectively. Task objects and information-bearing regions are placed at least $20^\circ$ from either angular boundary. In two-level tasks, the upper sector may be offset by $20^\circ$ relative to the lower sector, forcing the robot to actively switch viewing regions. 
An object is considered visible when at least 40\% of its projected bounding-box area lies
within the head-camera image and remains unoccluded by other scene elements, as determined
through depth-aware visibility checking. Each scene additionally contains a randomized number
of static distractors on the lower level.

\subsection{Benchmark Task Inventory and Subtask Semantics}
\label{app:task-inventory}

Table~\ref{tab:benchmark-task-inventory} specifies the benchmark at two
complementary semantic levels.

\begingroup
\scriptsize
\setlength{\tabcolsep}{3pt}
\setlength{\LTleft}{0pt}
\setlength{\LTright}{\fill}
\renewcommand{\arraystretch}{1.12}

\begin{longtable}{@{}
    >{\raggedright\arraybackslash}p{0.22\textwidth}
    >{\raggedright\arraybackslash}p{0.24\textwidth}
    >{\raggedright\arraybackslash}p{\dimexpr0.54\textwidth-4\tabcolsep\relax}@{}}
\caption{Benchmark tasks, instructions, and subtask definitions.}
\label{tab:benchmark-task-inventory}\\
\toprule
Task & Instruction & Subtasks \\
\midrule
\endfirsthead

\multicolumn{3}{c}{\tablename~\thetable\ (continued)}\\
\toprule
Task & Instruction & Subtasks \\
\midrule
\endhead

\midrule
\multicolumn{3}{r}{\footnotesize Continued on next page}\\
\endfoot

\bottomrule
\endlastfoot

\path{beat_block_hammer_rotate_view}
& Grab hammer using the left arm and hit the block.
& \textbf{T1} \path{pick_hammer}: locate and grasp the hammer.\newline
  \textbf{T2} \path{hammer_block}: locate the block and strike it with the carried hammer. \\
\addlinespace[2pt]

\path{blocks_ranking_rgb_fan_double}
& Arrange the red, green, and blue blocks from left to right.
& \textbf{T1} \path{pick_green_block}: locate and grasp the green block.\newline
  \textbf{T2} \path{place_green_block_right_of_red}: place the green block to the right of the red block.\newline
  \textbf{T3} \path{pick_blue_block}: locate and grasp the blue block.\newline
  \textbf{T4} \path{place_blue_block_right_of_green}: place the blue block to the right of the green block. \\
\addlinespace[2pt]

\path{blocks_ranking_rgb_rotate_view}
& Arrange the red, green, and blue blocks from left to right.
& \textbf{T1} \path{pick_green_block}: locate and grasp the green block.\newline
  \textbf{T2} \path{place_green_block_right_of_red}: place the green block to the right of the red block.\newline
  \textbf{T3} \path{pick_blue_block}: locate and grasp the blue block.\newline
  \textbf{T4} \path{place_blue_block_right_of_green}: place the blue block to the right of the green block. \\
\addlinespace[2pt]

\path{blocks_ranking_size_fan_double}
& Arrange the large, medium, and small blocks from left to right.
& \textbf{T1} \path{pick_medium_block}: locate and grasp the medium block.\newline
  \textbf{T2} \path{place_medium_block_right_of_large}: place the medium block to the right of the large block.\newline
  \textbf{T3} \path{pick_small_block}: locate and grasp the small block.\newline
  \textbf{T4} \path{place_small_block_right_of_medium}: place the small block to the right of the medium block. \\
\addlinespace[2pt]

\path{blocks_ranking_size_rotate_view}
& Arrange the large, medium, and small blocks from left to right.
& \textbf{T1} \path{pick_medium_block}: locate and grasp the medium block.\newline
  \textbf{T2} \path{place_medium_block_right_of_large}: place the medium block to the right of the large block.\newline
  \textbf{T3} \path{pick_small_block}: locate and grasp the small block.\newline
  \textbf{T4} \path{place_small_block_right_of_medium}: relocate the medium block and place the small block to its right. \\
\addlinespace[2pt]

\path{check_block_color}
& Inspect the gray block's hidden color and place it on the matching pad.
& \textbf{T1} \path{pick_target_block}: locate and pick up the gray target block.\newline
  \textbf{T2} \path{inspect_target_block_backside}: reveal and identify the block's backside color.\newline
  \textbf{T3} \path{restore_target_block_pose}: restore the block to a placement-ready pose.\newline
  \textbf{T4} \path{place_target_block_on_matching_pad}: localize the matching pad and place the block on it. \\
\addlinespace[2pt]

\path{check_cola_color}
& Inspect the can's hidden color and place it in the matching area.
& \textbf{T1} \path{pick_cola_can}: pick up the can.\newline
  \textbf{T2} \path{inspect_cola_backside_color}: expose and read the backside color label.\newline
  \textbf{T3} \path{restore_cola_after_inspection}: restore the can to a placement-ready pose.\newline
  \textbf{T4} \path{place_can_on_matching_color_area}: select the matching area and place the can. \\
\addlinespace[2pt]

\path{check_cola_date}
& Inspect the production date and sort the can by expiry status.
& \textbf{T1} \path{pick_cola_can}: pick up the can.\newline
  \textbf{T2} \path{inspect_cola_backside_date}: expose and read the production-date label.\newline
  \textbf{T3} \path{restore_cola_after_inspection}: restore the can to a placement-ready pose.\newline
  \textbf{T4} \path{place_cola_by_expiry}: infer expiry status, localize the corresponding area, and place the can. \\
\addlinespace[2pt]

\path{click_bell_rotate_view}
& Press the bell's top center using the left arm.
& \textbf{T1} \path{_click_bell}: locate the bell and press its top center. \\
\addlinespace[2pt]

\path{count_color_kinds_press_button}
& Count the distinct block colors and press the matching button.
& \textbf{T1} \path{count_unique_block_colors}: scan all blocks and infer the number of distinct colors.\newline
  \textbf{T2} \path{press_matching_number_button}: localize and press the button corresponding to the inferred count. \\
\addlinespace[2pt]

\path{count_random_object_press_button}
& Count the instructed objects and press the matching button.
& \textbf{T1} \path{count_random_target_objects}: scan the scene and count instances of the instructed category.\newline
  \textbf{T2} \path{press_matching_number_button}: localize and press the button corresponding to the inferred count. \\
\addlinespace[2pt]

\path{count_target_press_button}
& Count the target blocks and press the matching button.
& \textbf{T1} \path{count_green_target_blocks}: scan targets and distractors and infer the green-block count.\newline
  \textbf{T2} \path{press_matching_number_button}: localize and press the button corresponding to the inferred count. \\
\addlinespace[2pt]

\path{match_backside_two_blocks}
& Inspect two gray blocks and place each on its matching pad.
& \textbf{T1/T4} \path{pick_a_grey_block}: locate and pick the next gray block.\newline
  \textbf{T2/T5} \path{inspect_a_grey_block_backside_color}: expose and infer the carried block's backside color.\newline
  \textbf{T3/T6} \path{place_a_grey_block_on_matching_pad}: localize the matching pad and place the block. \\
\addlinespace[2pt]

\path{move_pillbottle_pad_rotate_view}
& Move the pill bottle onto the pad.
& \textbf{T1} \path{pick_pillbottle}: locate and pick up the pill bottle.\newline
  \textbf{T2} \path{place_pillbottle_on_pad}: localize the pad and place the carried bottle on it. \\
\addlinespace[2pt]

\path{move_stapler_pad_rotate_view}
& Move the stapler onto the black mat.
& \textbf{T1} \path{pick_stapler}: locate and pick up the stapler.\newline
  \textbf{T2} \path{place_stapler_on_pad}: localize the instructed mat and place the stapler on it. \\
\addlinespace[2pt]

\path{place_a2b_left_rotate_view}
& Place the Rubik's cube to the left of the wooden block.
& \textbf{T1} \path{pick_object_A}: locate and pick up source object A.\newline
  \textbf{T2} \path{place_A_left_of_B}: localize reference object B and place A to its left. \\
\addlinespace[2pt]

\path{place_a2b_right_rotate_view}
& Place the bell to the right of the coffee box.
& \textbf{T1} \path{pick_object_A}: locate and pick up source object A.\newline
  \textbf{T2} \path{place_A_right_of_B}: localize reference object B and place A to its right. \\
\addlinespace[2pt]

\path{place_cans_plasticbox_rotate_view}
& Place both cans into the plastic box.
& \textbf{T1} \path{pick_first_can}: locate and pick up the first can.\newline
  \textbf{T2} \path{place_first_can_in_box}: localize the box and place the first can inside.\newline
  \textbf{T3} \path{pick_second_can}: locate and pick up the second can.\newline
  \textbf{T4} \path{place_second_can_in_box}: relocate the box and place the second can inside. \\
\addlinespace[2pt]

\path{place_container_plate_rotate_view}
& Place the container on the plate.
& \textbf{T1} \path{pick_container}: locate and grasp the container.\newline
  \textbf{T2} \path{place_container_on_plate}: localize the plate and place the carried container on it. \\
\addlinespace[2pt]

\path{place_empty_cup_rotate_view}
& Place the empty cup on the coaster.
& \textbf{T1} \path{pick_cup}: locate and grasp the cup.\newline
  \textbf{T2} \path{place_cup_on_coaster}: localize the coaster and place the carried cup on it. \\
\addlinespace[2pt]

\path{place_fan_rotate_view}
& Place the fan on the silver mat facing the robot.
& \textbf{T1} \path{pick_fan}: locate and grasp the fan.\newline
  \textbf{T2} \path{place_fan_on_pad}: localize the instructed mat and place the fan with the required orientation. \\
\addlinespace[2pt]

\path{place_mouse_pad_rotate_view}
& Place the mouse on the cyan mat.
& \textbf{T1} \path{pick_mouse}: locate and grasp the mouse.\newline
  \textbf{T2} \path{place_mouse_on_pad}: localize the instructed pad and place the mouse on it. \\
\addlinespace[2pt]

\path{place_object_basket_fan_double}
& Place the target object into the bread basket.
& \textbf{T1} \path{pick_object}: locate and grasp the episode-specific target object.\newline
  \textbf{T2} \path{place_object_into_basket}: localize the bread basket and place the carried object inside. \\
\addlinespace[2pt]

\path{place_object_scale_rotate_view}
& Place the target object on the electronic scale.
& \textbf{T1} \path{pick_object}: locate and grasp the episode-specific target object.\newline
  \textbf{T2} \path{place_object_on_scale}: localize the scale and place the carried object on it. \\
\addlinespace[2pt]

\path{place_object_stand_rotate_view}
& Place the Rubik's cube on the display stand.
& \textbf{T1} \path{pick_object}: locate and grasp the target object.\newline
  \textbf{T2} \path{place_object_on_stand}: localize the display stand and place the carried object on it. \\
\addlinespace[2pt]

\path{place_shoe_rotate_view}
& Place the shoe on the mat.
& \textbf{T1} \path{pick_shoe}: locate and grasp the shoe.\newline
  \textbf{T2} \path{place_shoe_on_mat}: localize the mat and place the carried shoe on it. \\
\addlinespace[2pt]

\path{press_stapler_rotate_view}
& Firmly press the stapler.
& \textbf{T1} \path{_press_stapler}: localize and press the stapler. \\
\addlinespace[2pt]

\path{put_block_on_upper_easy}
& Place the block into the upper-level plate.
& \textbf{T1} \path{pick_a_block_0}: locate and pick up the block.\newline
  \textbf{T2} \path{place_a_block_0_into_plate}: localize the upper-level plate and place the block inside. \\
\addlinespace[2pt]

\path{put_block_on_upper_hard}
& Place all blocks into the plate.
& \textbf{T1} \path{pick_a_block_0}: select and pick up the first remaining block.\newline
  \textbf{T2} \path{place_a_block_0_into_plate}: localize the plate and place the first block inside.\newline
  \textbf{T3} \path{pick_a_block_1}: locate and pick up the remaining block.\newline
  \textbf{T4} \path{place_a_block_1_into_plate}: relocate the plate and place the second block inside. \\
\addlinespace[2pt]

\path{rank_backside_rgb_blocks}
& Inspect three gray blocks and arrange them in red--green--blue order.
& \textbf{T1/T4/T7} \path{pick_a_grey_block}: locate and pick the next gray block.\newline
  \textbf{T2/T5/T8} \path{inspect_a_grey_block_backside_color}: expose and infer its hidden backside color.\newline
  \textbf{T3/T6/T9} \path{place_a_grey_block_on_rgb_rank_pad}: localize the color-implied rank pad and place the block. \\
\addlinespace[2pt]

\path{shake_bottle_horizontally_rotate_view}
& Pick up the bottle and shake it horizontally.
& \textbf{T1} \path{_shake_bottle_horizontally}: locate and grasp the bottle, then execute horizontal shaking. \\
\addlinespace[2pt]

\path{shake_bottle_rotate_view}
& Pick up the bottle and shake it.
& \textbf{T1} \path{_shake_bottle}: locate and grasp the bottle, then execute the shaking motion. \\
\addlinespace[2pt]

\path{stack_blocks_two_rotate_view}
& Stack the green block on the red block.
& \textbf{T1} \path{pick_green_block}: locate and grasp the green block.\newline
  \textbf{T2} \path{stack_green_on_red}: localize the red block and stack the carried green block on it. \\
\addlinespace[2pt]

\path{stamp_seal_rotate_view}
& Move the seal to the beige pad and stamp it.
& \textbf{T1} \path{pick_seal}: locate and grasp the seal.\newline
  \textbf{T2} \path{stamp_target_pad}: localize the instructed pad and execute the stamping action. \\
\addlinespace[2pt]

\path{turn_switch_rotate_view}
& Locate and activate the switch.
& \textbf{T1} \path{_turn_switch}: localize and trigger the switch. \\

\end{longtable}

\endgroup

\subsection{ID/OOD Evaluation Protocol}
\label{app:evaluation-protocol}

Both the in-distribution (ID) and out-of-distribution (OOD) settings randomize task-object poses, object attributes, and clutter layouts while sharing the same workspace geometry, camera configuration, and success criteria.

The OOD setting introduces stronger visual distribution shifts through randomized backgrounds, tabletop textures, illumination, and object distributions, with clutter potentially including target objects from other tasks. Experimental results reveal a clear ID/OOD performance gap, indicating that this configuration provides a sufficiently challenging test of model generalization.

\subsection{History-Keyframe Definitions}
\label{app:keyframe-definitions}

Because there is no standard representation for robot visual history, we compare three complementary keyframe definitions: semantic Subtask Keyframes, motion-driven Motion Keyframes, and action-aligned Chunk Keyframes. For a trajectory of length $T$,
\begin{equation}
\tau=\{(o_t,a_t,s_t,x_t)\}_{t=0}^{T-1},
\end{equation}
where $o_t$ is the visual observation, $a_t$ is the action, $s_t$ is the discrete subtask label, and $x_t$ is the robot state. For each keyframe type
$z\in{\mathrm{sub},\mathrm{mot},\mathrm{chunk}}$, let $\mathcal K^z$ denote the retained frame set and
$k_t^z=\mathbb 1[t\in\mathcal K^z]$ its binary indicator. 

\paragraph{Subtask Keyframes.}
The first trajectory frame and the first frame following each change in the discrete subtask label are retained:
\begin{equation}
\mathcal K^{\mathrm{sub}}
=\{0\}\cup
\{t\mid s_t\neq s_{t-1},\ 1\leq t<T\}.
\label{eq:subtask-keyframes}
\end{equation}
For example, when a task transitions from ``search for and grasp
the hammer'' to ``locate and strike the target block,'' the first
frame of the latter subtask is retained. This definition captures high-level task progress and semantic phase transitions.

\paragraph{Motion Keyframes.}
Motion Keyframes retain meaningful motion-state boundaries rather than
uniformly sampled frames. Let $\mathcal{E}(x_{\leq t})$ denote the set of
events causally confirmed by time $t$. The default events include the
gripper entering or leaving a stable open or closed state, as well as the
onset or termination of head-pitch or torso-yaw rotation. The retained set is
\begin{equation}
\mathcal{K}^{\mathrm{mot}}
=
\operatorname{Dedup}_{\delta}\!\left(
\{0\}
\cup
\left\{
t \mid \mathcal{E}(x_{\leq t}) \neq \varnothing
\right\}
\right),
\qquad \delta=5.
\label{eq:motion_keyframes}
\end{equation}

A gripper is considered closed at values no greater than $0.2$ and open
at values no less than $0.8$; the state must persist for two frames before
confirmation. Head pitch or torso yaw is considered rotating when its
absolute frame-to-frame change exceeds $0.005\,\mathrm{rad}$. Rotation must
persist for four frames, and interruptions of at most five frames are merged.

We also provide end-effector-motion event detector. 
For arm $a\in\{\mathrm{L},\mathrm{R}\}$, let
$\mathbf{p}^{a}_{t}\in\mathbb{R}^{3}$ denote the end-effector position.
Its Cartesian velocity and speed are defined as
\begin{equation}
\mathbf{v}^{a}_{t}
=
\frac{\mathbf{p}^{a}_{t}-\mathbf{p}^{a}_{t-1}}{\Delta t},
\qquad
s^{a}_{t}
=
\left\|\mathbf{v}^{a}_{t}\right\|_{2}.
\end{equation}

We detect two types of end-effector events. A turning event is triggered
when the angle between two consecutive velocity vectors exceeds
$40^\circ$:
\begin{equation}
\alpha^{a}_{t}
=
\arccos\!\left(
\frac{
\left\langle\mathbf{v}^{a}_{t-1},\mathbf{v}^{a}_{t}\right\rangle
}{
s^{a}_{t-1}s^{a}_{t}
}
\right)
>40^\circ,
\end{equation}
where both $s^{a}_{t-1}$ and $s^{a}_{t}$ must be greater than
$\epsilon_{\mathrm{stop}}=0.001$.

A stopping event is triggered when the end effector changes from moving
to nearly stationary:
\begin{equation}
s^{a}_{t-1}>\epsilon_{\mathrm{stop}},
\qquad
s^{a}_{t}\leq\epsilon_{\mathrm{stop}},
\qquad
\epsilon_{\mathrm{stop}}=0.001.
\end{equation}

The resulting eef event set is
\begin{equation}
\mathcal{E}^{\mathrm{eef}}(x_{\leq t})
=
\left\{
(a,\mathrm{turn},t)
\mid
\alpha^{a}_{t}>40^\circ
\right\}
\cup
\left\{
(a,\mathrm{stop},t)
\mid
s^{a}_{t-1}>0.001,\,
s^{a}_{t}\leq0.001
\right\}.
\end{equation}

\paragraph{Chunk Keyframes.}
Chunk Keyframes provide a uniform temporal baseline aligned with policy action chunks rather than motion changes. With a fixed chunk length of $C=16$,
\begin{equation}
\mathcal K^{\mathrm{chunk}}
={nC\mid n\in\mathbb N_0,\ nC<T},
\qquad
k_t^{\mathrm{chunk}}
=\mathbb 1[t\bmod 16=0].
\label{eq:chunk-keyframes}
\end{equation}
Each Chunk Keyframe corresponds to a timestep at which the policy receives a new observation, predicts the next action chunk, and replans. 

\paragraph{Unified Input Construction.}
For keyframe type $z$, the visual memory at time $t$ consists of the most recent $H$ retained observations strictly preceding the current frame:
\begin{equation} 
M_t^z
=\operatorname{Recent}_{H}
\bigl(\{o_i\mid i<t,\ i\in\mathcal K^z\}\bigr).
\qquad 
a_t\sim\pi_\theta(a_t\mid M_t^z,o_t,g), 
\label{eq:state-memory} 
\end{equation} 
where $g$ denotes the task instruction. 

\begin{equation}  % State-history input sequence.
X_t^z
=
\operatorname{Recent}_{H}
\bigl(\{x_i\mid i<t,\ i\in\mathcal K^z\}\bigr)
\mathbin{\Vert} x_t,
\end{equation} 
The state-conditioned policy is 
\begin{equation} 
a_t\sim\pi_\theta(a_t\mid M_t^z,o_t,g,X_t^z).  % State-conditioned policy.
\end{equation}

We use $H=12$, corresponding to at most 12 historical keyframes and one current frame. 

\section{Model Training and Evaluation Details}
\label{app:training-evaluation}

\subsection{Evaluation Resources}
\label{app:unified-evaluation}

\paragraph{Deterministic seed generation and storage.}
The evaluation environment is configured as follows:
\begin{itemize}
    \item Ubuntu 22.04;
    \item Python 3.10;
    \item CUDA 12.1;
    \item PyTorch 2.4.1.
\end{itemize}

Before evaluating the models, we perform a simulator-only run without loading
any policy model to collect the random seeds for all evaluation tasks. The
resulting seed manifest is shared across all models, ensuring that every model
is evaluated using the same set of random seeds.

\paragraph{RTX-4090 reproduction.}
Table~\ref{tab:single-4090-evaluation} reports the reference configuration for
evaluating an OFT policy with a 12-frame memory and 16-step action chunks on a
single NVIDIA RTX 4090. For $N$ NVIDIA RTX 4090 GPUs, the same seed manifest can be evaluated using
$N$ independent evaluation workers in parallel.

\begin{reporttable}[0.85\linewidth]
\caption{Reference evaluation profile on a single NVIDIA RTX 4090.}
\label{tab:single-4090-evaluation}
    \centering
    \small
    \setlength{\tabcolsep}{5pt}
    \renewcommand{\arraystretch}{1.12}

    \begin{tabular*}{\linewidth}{@{\extracolsep{\fill}}lc}
        \toprule
        Configuration & Value \\
        \midrule
        GPU & NVIDIA RTX 4090, 24 GB \\
        Peak GPU memory & 21.8 GB \\
        Peak host memory & 14.6 GB \\
        Mean throughput & 93.3 episodes/hour \\
        ID episodes & 1750 \\
        OOD episodes & 1750 \\
        Complete ID+OOD runtime & 37.50 hours \\
        \bottomrule
    \end{tabular*}
\end{reporttable}

\subsection{External Baseline Training Settings}
\noindent\textbf{$\pi_{0.5}$.}
We fine-tune $\pi_{0.5}$ using the current head-camera image and its aligned 18-dimensional proprioceptive state. Wrist-camera input slots are masked throughout training, while all other optimization settings follow the default \textsc{OpenPI} configuration. At evaluation time, each action chunk is generated using five flow-matching sampling steps.

\noindent\textbf{HiF-VLA.}
Following the original configuration, we fine-tune OpenVLA-7B using LoRA with rank 32 and zero dropout. Each training sample contains the current head-camera image, its aligned 18-dimensional proprioceptive state, and eight motion-vector entries spanning timesteps $t-7$ through $t$. Images are resized to $224{\times}224$ and normalized for the fused DINOv2/SigLIP visual backbone. Under quantile normalization, the model directly regresses an eight-step action chunk, with each step represented as an 18-dimensional absolute control vector. At inference time, the corresponding inverse normalization is applied, and the two gripper channels are clipped to $[0,1]$.

\noindent\textbf{MemER.}
The MemER high-level policy is fine-tuned from Qwen3-VL-4B-Instruct, while $\pi_{0.5}$ serves as the low-level controller. We otherwise retain the original training pipeline. The high-level policy receives up to eight selected memory frames, followed by up to eight recent head-camera frames sampled at a stride of five. It jointly predicts the current subtask and the location of the next keyframe.

\noindent\textbf{Fast-WAM.}
We adapt Fast-WAM to a single head-camera stream and initialize it from the Wan2.2-5B video DiT, T5 text encoder, and video VAE. Each training sample contains the current head-camera image, its aligned 18-dimensional proprioceptive state, future head-camera frames, and a 32-step action chunk of 18-dimensional absolute controls. The current frame serves as a clean visual anchor, whereas future frames are used only for the auxiliary video flow-matching objective; the action branch is causally prevented from accessing future visual tokens. Training follows the original joint action--video flow-matching objective. At inference time, the future-video branch is removed, and the action chunk is generated from the current observation using ten flow-matching steps.

\noindent\textbf{MemoryVLA.}
We retain the original Prismatic-7B vision-language backbone, perceptual--cognitive memory bank, and diffusion action expert. The head-camera image is encoded by the fused DINOv2/SigLIP visual backbone, while the aligned 18-dimensional proprioceptive state is linearly projected as an additional condition for the action expert. The memory bank stores at most 16 perceptual--cognitive entries and is updated online through historical retrieval, gated fusion, and adjacent-entry merging. We replace the original action head with an 18-dimensional absolute-control head that predicts a 16-step action chunk. All other training settings remain unchanged. Inference uses ten DDIM sampling steps with a classifier-free guidance weight of 1.5.

\noindent\textbf{SaPaVe.}
We retain the Eagle-2 camera adapter, MapAnything spatial encoder, and decoupled diffusion action heads. The original camera-action branch is adapted to a two-dimensional active-view controller that predicts torso yaw and the single-DoF head pitch, while the manipulation branch predicts 14 arm-joint commands and two gripper commands. Training follows the original two-stage procedure. In the first stage, only the LoRA camera adapter and active-view decoder are optimized on search-and-focus segments, with the spatial encoder and manipulation decoder frozen. In the second stage, the camera adapter is frozen and both action branches are jointly trained on complete demonstrations. Their outputs are concatenated along the action dimension to form an 18-dimensional action chunk.

\subsection{Visualization of Memoryless-Model Failures}
\label{app:memoryless-failures}
\begin{figure}[!tp]
    \centering
    \includegraphics[width=\linewidth]{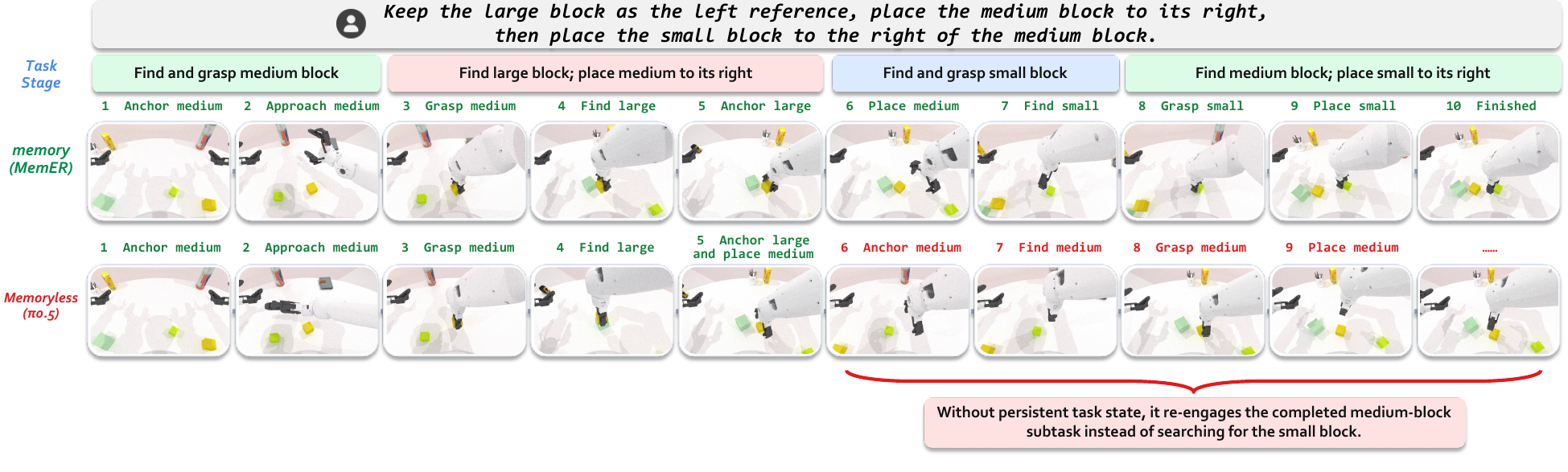}
    \caption{Paired rollouts of the memory-aware policy MemER (top) and the memoryless policy $\pi_{0.5}$ (bottom) on the block-arrangement task.}
    \label{fig:memoryless-blocks}
\end{figure}

\begin{figure}[!tp]
    \centering
    \includegraphics[width=\linewidth]{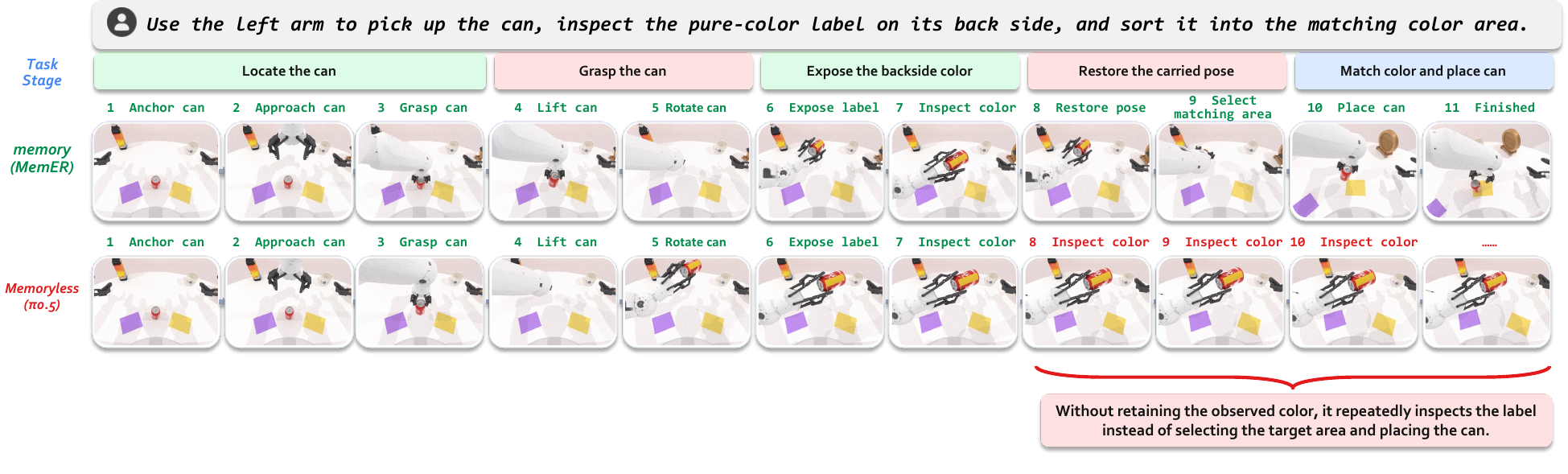}
    \caption{Paired rollouts of the memory-aware policy MemER (top) and the memoryless policy $\pi_{0.5}$ (bottom) on the hidden-color sorting task.}
    \label{fig:memoryless-cola}
\end{figure}

We use $\pi_{0.5}$ as a representative memoryless policy and compare it with MemER to analyze failures caused by the lack of persistent historical information. Figures~\ref{fig:memoryless-blocks} and~\ref{fig:memoryless-cola} present paired rollouts: both policies receive the same instruction and start from scene configurations generated with the same random seed. In these tasks, the next action depends not only on the current visual observation, but also on completed subgoals and evidence acquired from previous viewpoints. MemER is able to preserve this execution context across manipulation steps and viewpoint changes.

In the block-arrangement task, MemER remembers that the medium block has already been placed relative to the large block and therefore shifts its search toward the small block. In contrast, $\pi_{0.5}$ re-enters the visually plausible medium-block subtask and repeatedly manipulates the same object. In the hidden-color sorting task, MemER retains the backside color observed during inspection after the object is reoriented and uses this information to select the matching target area. By contrast, $\pi_{0.5}$ fails to retain the inspection result as evidence for subsequent decisions, repeatedly returning to the inspection stage without completing the placement. These failures show that losing execution history and previously acquired evidence can reduce long-horizon manipulation to locally repetitive behavior.

\subsection{Internal Baseline Details}

This section provides the implementation and optimization details for the 13
internal baseline policies.  All variants use the same training
set and the same head-camera observation stream.  The comparison changes only
the action-generation mechanism, language-level auxiliary supervision,
temporal keyframe selection, memory length, or proprioceptive-state
conditioning.  Table~\ref{tab:inner_baseline_variants} gives the complete run
manifest, while Tables~\ref{tab:shared_training_hparams}
and~\ref{tab:head_specific_hparams} report the shared and head-specific
hyperparameters, respectively.

\subsubsection{Shared Training Protocol}

A run
took approximately 50 hours, corresponding to approximately 400 A800 GPU-hours
per model.  Runs with auxiliary
vision--language co-training draw one action batch and one language-supervision batch
at each step.

All parameters are optimized and we use AdamW with $\beta_1=0.9$, $\beta_2=0.95$,
$\epsilon=10^{-8}$, and weight decay $10^{-8}$.  The Qwen backbone and its
interface use a peak learning rate of $10^{-5}$, whereas the state encoder and
continuous action heads use $10^{-4}$. 

\begin{table}[!htb]
\caption{Shared optimization and data hyperparameters for all inner
  baselines. }
\label{tab:shared_training_hparams}
  \centering

  \small
  \begin{tabularx}{\linewidth}{@{}>{\raggedright\arraybackslash}X
      >{\raggedright\arraybackslash}X@{}}
    \toprule
    Setting & Value \\
    \midrule
    Hardware & $8\times$ NVIDIA A800, single node \\
    Training budget & $150{,}000$ optimizer steps; approximately 50 h per run \\
    Precision / distribution & bfloat16; DeepSpeed ZeRO-2; no CPU offload \\
    Batch size & 7 per GPU \\
    Gradient accumulation & 1 \\
    Optimizer & AdamW, $\beta=(0.9,0.95)$, $\epsilon=10^{-8}$ \\
    Weight decay / gradient clipping & $10^{-8}$ / global norm 1.0 \\
    Backbone learning rate & $10^{-5}$ \\
    State/action-head learning rate & $10^{-4}$ \\
    Schedule & 5,000-step linear warmup, then cosine decay \\
    Minimum learning rate & $5\times10^{-7}$ \\
    Random seed & 42 \\
    Action representation & 18-D absolute control; 16-step action chunk \\
    \bottomrule
  \end{tabularx}
\end{table}
\paragraph{Inputs and targets.} 
Each sample contains the current head-camera image and, when enabled, up to
$K\in\{6,12\}$ historical images.  Consequently, a 12-frame memory variant
receives at most 13 images (12 history frames plus the current frame).  For
\emph{action} history, frames are sampled backward at a stride of 16 environment
steps.  The no-history variant uses only the current image.  

The action target is a 16-step chunk of 18-D absolute controls.  State-enabled
models receive an aligned 18-D proprioceptive vector for every input image.  A
two-layer MLP with hidden width 1,024 projects each vector into one soft state
token; learned frame embeddings preserve temporal identity.

\paragraph{Language supervision.}
Every action branch is prompted with the full task instruction.  The second
field in an OFT run name controls only the auxiliary language branch:
\texttt{no} disables this branch, \texttt{instruction} supervises the
\texttt{<think>} target with the full task instruction, and
\texttt{subtask} supervises it with the annotated current subtask.  For
co-trained OFT and GR00T runs, the total loss is
\begin{equation}
  \mathcal{L}=\mathcal{L}_{\mathrm{action}}
  +0.1\,\mathcal{L}_{\mathrm{VLM}}.
  \label{eq:inner_cotrain_loss}
\end{equation}
The FAST policy places the same subtask-level reasoning target before the
discrete action tokens in a single autoregressive response and is therefore
trained with one token-level cross-entropy objective rather than the separate
loss in Eq.~\ref{eq:inner_cotrain_loss}.

\subsubsection{\mnm-VLA Model Variants}
\label{app:inner_model_variants}
\begin{table}[!htb]
\caption{Complete internal-baseline manifest. ``Think'' is the auxiliary
  language target, $K$ is the maximum number of history frames, and ``State''
  indicates per-image proprioceptive soft-token conditioning.}
\label{tab:inner_baseline_variants}
  \centering

  \scriptsize
  \setlength{\tabcolsep}{3.2pt}
  \begin{tabularx}{\linewidth}{@{}>{\raggedright\arraybackslash}p{0.29\linewidth}
      >{\raggedright\arraybackslash}p{0.11\linewidth}
      >{\raggedright\arraybackslash}p{0.12\linewidth}
      >{\raggedright\arraybackslash}p{0.14\linewidth}
      >{\centering\arraybackslash}p{0.035\linewidth}
      >{\centering\arraybackslash}p{0.055\linewidth}
      >{\raggedright\arraybackslash}X@{}}
    \toprule
    Run & Head & Think & History & $K$ & State & Action objective \\
    \midrule
    \path{oft_instruction_action_12_ws} & OFT & instruction & action & 12 & yes & continuous L1 \\
    \path{oft_no_action_12_ws} & OFT & none & action & 12 & yes & continuous L1 \\
    \path{oft_subtask_no_0_ws} & OFT & subtask & none & 0 & yes & continuous L1 \\
    \path{oft_subtask_action_12_wos} & OFT & subtask & action & 12 & no & continuous L1 \\
    \path{oft_subtask_action_12_ws} & OFT & subtask & action & 12 & yes & continuous L1 \\
    \path{oft_subtask_action_6_ws} & OFT & subtask & action & 6 & yes & continuous L1 \\
    \path{oft_subtask_motion_12_ws} & OFT & subtask & motion & 12 & yes & continuous L1 \\
    \path{oft_subtask_motion_6_ws} & OFT & subtask & motion & 6 & yes & continuous L1 \\
    \path{oft_subtask_subtask_12_ws} & OFT & subtask & subtask & 12 & yes & continuous L1 \\
    \path{oft_subtask_subtask_6_ws} & OFT & subtask & subtask & 6 & yes & continuous L1 \\
    \path{gr00t_subtask_action_12_ws} & GR00T & subtask & action & 12 & yes & flow matching \\
    \path{fast_subtask_action_12_ws} & FAST & subtask & action & 12 & yes & token cross-entropy \\ 
    \path{oft_subtask_planned_12_ws} & OFT & subtask & planned & 12 & yes & continuous L1 \\
    \bottomrule
  \end{tabularx}
\end{table}

Run names follow
\texttt{architecture\_think\_history\_memory\_state}.  Here \texttt{ws} and
\texttt{wos} denote with and without proprioceptive state.  The memory field is
the maximum number of \emph{historical} frames and excludes the current frame.
The 13 configurations in Table~\ref{tab:inner_baseline_variants} form controlled
comparisons along these axes.

\subsubsection{\mnm-FAST}

\texttt{fast\_subtask\_action\_12\_ws} uses the same 12-frame action history
and state conditioning, but discretizes the $16\times18$ action chunk with the
FAST tokenizer and predicts it autoregressively.  We initialize from
Qwen3-VL-2B-Instruct-Action, which extends Qwen3-VL-2B-Instruct with 2,048
pre-initialized robot-action tokens. The target sequence concatenates a subtask-supervised
\texttt{<think>} span and the FAST codes inside an \texttt{<action>} span, and
the complete response is optimized by token cross-entropy.

\begin{table}[!htb]
\caption{Action-head-specific settings. Shared entries such as the 18-D
  action space, 16-step horizon, optimizer, and training budget are reported in
  Table~\ref{tab:shared_training_hparams}.}
\label{tab:head_specific_hparams}
  \centering

  \small
  \begin{tabularx}{\linewidth}{@{}>{\raggedright\arraybackslash}p{0.15\linewidth}
      >{\raggedright\arraybackslash}p{0.25\linewidth}
      >{\raggedright\arraybackslash}X@{}}
    \toprule
    Family & Initialization & Head-specific configuration \\
    \midrule
    OFT & Qwen3-VL-2B-Instruct & 16 action-query positions; 2,048-D input, 4,096-D hidden MLPResNet; 2 residual blocks; L1 regression \\
    GR00T & Qwen3-VL-2B-Instruct & DiT-B, width 1,024, 16 layers; 32 vision tokens; 8 training noise repeats; 4 inference steps \\
    FAST & Qwen3-VL-2B-Instruct-Action & FAST tokenizer; 2,048 action tokens; autoregressive cross-entropy \\
    \bottomrule
  \end{tabularx}
\end{table}

\subsubsection{\mnm-Plan Architecture}

We design \mnm-Plan: A planner observes the global task
instruction, a bounded history, and the current image.  It autoregressively
predicts the current subtask and indices of frames worth retaining.  A separate
VLA executor is conditioned on that subtask, the retrieved memory
frames, the current image, and frame-aligned robot state, and predicts only the
continuous action chunk.  Thus, the planner decides \emph{what stage the task
is in} and \emph{what to remember}, while the VLA decides \emph{how to execute
the current stage}.

\paragraph{Training Details.} For a training sample at current trajectory frame $x$, let the stride be
$s=16$.  We first construct the full stride-aligned sequence
\begin{equation}
  \mathcal{S}_{\infty}(x)
  =\operatorname{sort}\{x-js\mid j\in\{0,1,2,\ldots\},\ x-js\geq0\}.
  \label{eq:planner_stride_sequence}
\end{equation}
The planner keeps only the most recent $K+1$ elements, where $K=12$ is the
maximum number of \emph{history} frames and the final element is always $x$.
It therefore sees at most 13 images.  Near the beginning of a trajectory the
sequence is naturally shorter; sampling stops before an index would become
negative.  Images are ordered from earliest to latest and are followed by the
global task prompt.  Under the Qwen chat template, this is serialized
conceptually as $N$
consecutive image items followed by the prompt text; the
\texttt{<image>} markers are not inserted as ordinary hand-written text tokens:
\begin{quote}
  \small\ttfamily\raggedright
  Your task is: ``\{instruction\}'' The input images are ordered from earliest
  to latest, and the last image is the current view. Please think about the
  current subtask and which frame(s) among the input images are worth
  remembering. Your response should be in the format of:
  <think>...</think>\allowbreak <subtask>...</subtask>\allowbreak
  <retrieval>...\\</retrieval>.
\end{quote}
The planner receives images and language but no robot state.  Its target has
exactly one \texttt{<subtask>} span.  We use FlashAttention-2, a maximum
language-model sequence length of 4,096 tokens, and dynamic image resizing
between 784 and 50,176 pixels:
\begin{quote}
  \small\raggedright
  \texttt{<think>Frames: \{num\_frames\} total
  (\{num\_history\} history + current). Now the}\\
  \texttt{subtask is
  ``\{subtask instruction\}''</think>}\allowbreak
  \texttt{<subtask>\{subtask instruction\}</subtask>}\allowbreak\\
  \texttt{<retrieval>[keyframe indices]</retrieval>}
\end{quote}
Only the current subtask instruction is passed as language conditioning to the
executor; the \texttt{<think>} span provides structured autoregressive
supervision, and \texttt{<retrieval>} controls visual memory.

\paragraph{Retrieval labels.}
Let $q$ be a previously observed subtask keyframe and let $f_i$ denote the
frame at zero-based position $i$ in the planner input.  Position $i$ is a
positive retrieval target exactly when
\begin{equation}
  q\leq f_i<q+s.
  \label{eq:planner_retrieval_window}
\end{equation}
Equivalently, $f_i$ must lie in the half-open window $[q,q+16)$.  We emit the
unique matching input positions in chronological order.  Frames outside all
such windows are not retrieval targets.

For example, when $x=89$, Eq.~\ref{eq:planner_stride_sequence} gives the input
frame indices
\begin{equation}
  [9,25,41,57,73,89].
\end{equation}
Suppose the observed subtask keyframes are $q=0$ and $q=45$.  Frame 9 lies in
$[0,16)$ and frame 57 lies in $[45,61)$, so their positions in the six-image
input are 0 and 3.  The complete target is
\begin{quote}
  \small\raggedright
  \texttt{<think>Frames: 6 total (5 history + current). Now the subtask is}\\
  \texttt{``\{subtask instruction\}''</think>}\allowbreak\\
  \texttt{<subtask>\{subtask instruction\}</subtask>}\allowbreak
  \texttt{<retrieval>[0, 3]</retrieval>}.
\end{quote}
The textual placeholders are filled with the subtask annotation for frame 89.

\paragraph{Executor Memory and Action Target.}

The executor uses the same stride-16 sequence but does \emph{not} apply the
planner's $K=12$ history cap.  During supervised VLA data construction, we
apply Eq.~\ref{eq:planner_retrieval_window} to all subtask keyframes and all
frames in $\mathcal{S}_{\infty}(x)$.  The selected memory frames are arranged
chronologically and followed by the current frame $x$; if $x$ is already the
latest selected memory frame, it is not duplicated.  Each image is paired with
its aligned 18-D state vector.

The executor
is prompted with the predicted current subtask rather than the global
instruction:
\begin{quote}
  \small\ttfamily\raggedright
  Your task is: ``\{subtask instruction\}'' The input images are ordered from
  earliest to latest, and the last image is the current view. Please think
  about the next\\ action.
\end{quote}
It uses the same OFT configuration described.  

\paragraph{Resource Efficiency.} Training \path{oft_subtask_planned_12_ws} reduces peak
  GPU memory usage by 46\% compared with the best dense-history model,
  \path{oft_subtask_action_12_ws}, decreasing from 78.9 GB to 42.6 GB.

% Required packages:
% \usepackage{booktabs}
% \usepackage{tabularx}
% \usepackage{array}

\section{Details of Real-World Experiments}
\label{app:real-world}

\subsection{Hardware and Software Setup}

\paragraph{Robot platform and observations.}
All real-world experiments are conducted on an Astribot S1. The platform
is equipped with two 7-DoF arms, two grippers, an actuated
torso, and an actuated head. The policy receives egocentric
RGB observations from the head-mounted camera, together with
the language instruction and robot proprioceptive state.

Consistent with the simulation setup, the policy predicts an
18-D action:
\begin{equation}
    \mathbf{a}_t =
    [\mathbf{a}^{L}_t,g^{L}_t,
     \mathbf{a}^{R}_t,g^{R}_t,
     a^{\mathrm{torso}}_t,a^{\mathrm{head}}_t],
\end{equation}
which jointly controls the two arms, two grippers, torso, and
head.

\paragraph{Calibration and execution.}
Before each evaluation session, we verify the camera
intrinsics, camera-to-head extrinsics, robot kinematics, and
the zero configurations of the head and torso. Policy
inference is performed on a dedicated GPU workstation
connected to the robot controller through a wired local
network. For every episode, we record RGB observations,
proprioceptive states, predicted actions, controller
feedback, and timestamps.

\paragraph{Capability taxonomy.}
We use four operational labels to characterize the active
perception capabilities required by the real-world tasks.

\begingroup
\small
\setlength{\tabcolsep}{5pt}
\setlength{\LTleft}{0pt}
\setlength{\LTright}{\fill}
\renewcommand{\arraystretch}{1.12}
\begin{longtable}{@{}
    >{\raggedright\arraybackslash}p{0.08\linewidth}
    >{\raggedright\arraybackslash}p{0.16\linewidth}
    >{\raggedright\arraybackslash}p{\dimexpr0.76\linewidth-4\tabcolsep\relax}@{}}
\caption{Operational definitions of the active perception
capabilities used in the real-world evaluation.}
\label{tab:real-world-capabilities}\\
\toprule
Symbol & Capability & Operational meaning \\
\midrule
\endfirsthead
\multicolumn{3}{c}{\tablename~\thetable\ (continued)}\\
\toprule
Symbol & Capability & Operational meaning \\
\midrule
\endhead
\midrule
\multicolumn{3}{r}{\footnotesize Continued on next page}\\
\endfoot
\bottomrule
\endlastfoot
S & Search &
The robot actively changes its head or torso viewpoint and
examines different workspace regions to locate task-relevant
objects, containers, or destination baskets. \\
\addlinespace[3pt]

T & Track &
The robot maintains visual alignment with a moving or
task-relevant target during execution, such as tracking a
moving recipient in a handover task. \\
\addlinespace[3pt]

I & Interact &
The robot physically manipulates containers, occluders, or
candidate objects to reveal information unavailable through
passive observation or to complete physical subtasks. \\
\addlinespace[3pt]

F & Focus &
The robot moves its camera, head, or a manipulated object to
obtain a clearer view of fine-grained visual attributes, such
as the size label of a T-shirt. \\
\end{longtable}
\endgroup

\paragraph{Real-world task suite.}
We construct eight real-world manipulation tasks and collect
640 successful robot trajectories for policy fine-tuning.
Table~\ref{tab:real-world-tasks} summarizes the task
instructions, required capabilities, and primary challenges.

\begingroup
\small
\setlength{\tabcolsep}{5pt}
\setlength{\LTleft}{0pt}
\setlength{\LTright}{\fill}
\renewcommand{\arraystretch}{1.12}
\begin{longtable}{@{}
    >{\raggedright\arraybackslash}p{0.05\linewidth}
    >{\raggedright\arraybackslash}p{0.26\linewidth}
    >{\raggedright\arraybackslash}p{0.13\linewidth}
    >{\raggedright\arraybackslash}p{\dimexpr0.56\linewidth-6\tabcolsep\relax}@{}}
\caption{Real-world task suite and required active perception
capabilities.}
\label{tab:real-world-tasks}\\
\toprule
ID & Task & Capability & Main challenge \\
\midrule
\endfirsthead
\multicolumn{4}{c}{\tablename~\thetable\ (continued)}\\
\toprule
ID & Task & Capability & Main challenge \\
\midrule
\endhead
\midrule
\multicolumn{4}{r}{\footnotesize Continued on next page}\\
\endfoot
\bottomrule
\endlastfoot
T1 &
Find the pen and place it in the basket &
S, I &
Search through a cloth bag until the pen is found, retrieve
it, locate the basket, and complete the placement. \\
\addlinespace[3pt]

T2 &
Find the peach and place it in the basket &
S, I &
Identify which drawer of a two-drawer cabinet contains the
peach, retrieve it, and place it in the basket. \\
\addlinespace[3pt]

T3 &
Find the screwdriver and place it in the basket &
S, I &
Search through cluttered objects, identify the screwdriver,
and transfer it to the basket. \\
\addlinespace[3pt]

T4 &
Find the XL-size T-shirt and place it in the basket &
S, I, F &
Inspect two visually similar T-shirts, identify the one with
the XL-size label, and place it in the basket. \\
\addlinespace[3pt]

T5 &
Find the grape and place it in the basket &
S, I &
Determine which drawer of a three-drawer cabinet contains the
grape, retrieve it, and complete the placement. \\
\addlinespace[3pt]

T6 &
Find the glue stick and place it in the basket &
S, I &
Search a cloth bag or drawer for the glue stick, retrieve it,
and transfer it to the basket. \\
\addlinespace[3pt]

T7 &
Find the banana and place it in the basket &
S, I &
Search a three-drawer cabinet containing distractor fruits,
identify the banana, and place it in the basket. \\
\addlinespace[3pt]

T8 &
Dynamic multi-person tracking and object handover &
S, I, T &
Continuously distinguish and track members of the red team
while both teams are moving, receive a bottle from red-team
player 1, and deliver it to red-team player 3. \\
\end{longtable}
\endgroup

\paragraph{Reset and evaluation protocol.}
Each method is evaluated for 20 trials on each of the eight
tasks. After every trial, the robot returns to a fixed home
configuration. An operator then reconstructs the predefined
scene, including the target, source container, distractors,
occluders, basket, background, illumination, and initial
robot pose.

\subsection{Randomized Perturbations}

\paragraph{Per-episode randomization.}
We introduce task-preserving perturbations to prevent the
policies from exploiting fixed object coordinates, camera
poses, or memorized action sequences. For each task, we
generate 20 configurations in advance, and replay every
configuration for all evaluated methods.

\begingroup
\small
\setlength{\tabcolsep}{5pt}
\setlength{\LTleft}{0pt}
\setlength{\LTright}{\fill}
\renewcommand{\arraystretch}{1.12}
\begin{longtable}{@{}
    >{\raggedright\arraybackslash}p{0.20\linewidth}
    >{\raggedright\arraybackslash}p{0.32\linewidth}
    >{\raggedright\arraybackslash}p{\dimexpr0.48\linewidth-4\tabcolsep\relax}@{}}
\caption{Per-episode randomization used in the real-world
evaluation. All pose offsets are defined relative to a
task-specific nominal configuration.}
\label{tab:real-world-randomization}\\
\toprule
Variable & Sampling range & Validity constraint \\
\midrule
\endfirsthead
\multicolumn{3}{c}{\tablename~\thetable\ (continued)}\\
\toprule
Variable & Sampling range & Validity constraint \\
\midrule
\endhead
\midrule
\multicolumn{3}{r}{\footnotesize Continued on next page}\\
\endfoot
\bottomrule
\endlastfoot
Target position &
\(\Delta x,\Delta y\sim\mathcal{U}(-5,5)\) cm &
The target remains reachable but cannot be grasped directly
from the initial configuration. \\
\addlinespace[3pt]

Target orientation &
\(\Delta\psi\sim\mathcal{U}(-20^\circ,20^\circ)\) &
The range is extended to \(\pm45^\circ\) for elongated tools
and flexible objects. \\
\addlinespace[3pt]

Source-region pose &
Position within \(\pm5\) cm; yaw within \(\pm15^\circ\) &
The container or occluder arrangement remains physically
operable. \\
\addlinespace[3pt]

Basket pose &
Position within \(\pm10\) cm; yaw within \(\pm20^\circ\) &
The basket remains spatially separated from the source
region. \\
\addlinespace[3pt]

Distractors &
\(N_{\mathrm{dist}}\sim\operatorname{Unif}\{3,4,5,6\}\) &
Distractor identities are sampled from a task-irrelevant
object pool. \\
\addlinespace[3pt]

Occluders &
\(N_{\mathrm{occ}}\sim\operatorname{Unif}\{1,2,3\}\) &
The target is initially occluded but can be revealed through
feasible interaction. \\
\addlinespace[3pt]

Head initialization &
Pitch within \(\pm10^\circ\) &
The sampled pose is clipped to the safe joint range. \\
\addlinespace[3pt]

Torso initialization &
Yaw within \(\pm10^\circ\) &
Both arms begin from the same safe home configuration. \\
\addlinespace[3pt]

Illumination &
Intensity scale \(\sim\mathcal{U}(0.7,1.3)\); light azimuth
within \(\pm25^\circ\) &
The resulting image must not be severely underexposed or
saturated. \\*[3pt]

Background &
Uniform sampling from a predefined set &
The same physical background set is used for every method. \\*[3pt]

T-shirt configuration &
Randomized candidate order, overlap, orientation, and label
pose &
The size label is not reliably readable initially but becomes
observable after feasible manipulation. \\
\end{longtable}
\endgroup

\paragraph{Experimental results.}
Table~\ref{tab:real-world-results} reports the success rates
of the evaluated methods across the eight real-world tasks.
\begin{table}[!htb]
\caption{Success rates (\%) on the real-world task suite.}
\label{tab:real-world-results}
    \centering
    \small
    \renewcommand{\arraystretch}{1.2}
    \setlength{\tabcolsep}{7pt}

    \begin{tabularx}{\linewidth}{
        >{\raggedright\arraybackslash}Xcccc}
        \toprule
        \textbf{Task}
        & \textbf{\(\pi_{0.5}\)}
        & \textbf{SaPaVe}
        & \textbf{MemER}
        & \textbf{\mnm-OFT} \\
        \midrule

        Locate the pen and place it in the basket.
        & 45 & 55 & 65 & 75 \\

        Retrieve the peach and transfer it to the basket.
        & 30 & 35 & 50 & 70 \\

        Find the screwdriver and deposit it in the basket.
        & 25 & 15 & 30 & 55 \\

        Select the XL-size T-shirt and place it in the basket.
        & 0 & 5 & 10 & 40 \\

        Retrieve the grape and put it in the basket.
        & 35 & 40 & 45 & 70 \\

        Locate the glue stick and move it to the basket.
        & 25 & 25 & 40 & 65 \\

        Find the banana and complete the placement.
        & 25 & 15 & 25 & 50 \\

        Track the relay game with moving red and blue teams.
        & 5 & 5 & 10 & 20 \\

        \midrule
        \textbf{Average}
        & \textbf{23.75}
        & \textbf{24.38}
        & \textbf{34.38}
        & \textbf{55.63} \\

        \bottomrule
    \end{tabularx}
\end{table}

\section{Further Discussion}\label{sec}

\paragraph{Why do we introduce an OOD setting, and is the current OOD protocol sufficient?}Recent studies have increasingly recognized that many existing robotic manipulation benchmarks essentially evaluate data fitting within a single distribution~\citep{fei2025liberoplus,chen2026robodojo,yuan2026qwen}.For example, benchmarks such as LIBERO often mix clean and randomized scenes within the same overall data distribution. While this setting is suitable for evaluating a model's ability to learn tasks within a given distribution, it is inadequate for assessing generalization, because it cannot reveal how the model performs when the scene distribution changes.

We therefore provide two complementary evaluation protocols. The ID setting measures task learning within the training distribution, whereas the OOD setting evaluates generalization to changed scenes. Our OOD scenes include unseen backgrounds, illumination changes, a broader and more challenging object asset pool, and target categories from other tasks appearing as distractors in the current task. These changes are sufficient to test whether a model relies on visual patterns and distractor configurations specific to the training scenes. The substantial ID--OOD performance gaps observed across most models further indicate that the protocol is sufficiently discriminative, at least to a meaningful extent, for comparing their generalization ability.

\paragraph{Are external baselines compared under fair input conditions?}Existing models require different forms of input, and this difference is even more pronounced for memory-based methods. Since research on robotic memory has not yet converged on a unified formulation, different methods rely on different memory representations and input interfaces.

As a benchmark, we therefore aim to provide training data that can support the input requirements of diverse methods. \mnm-Sim records rich annotations so that each model can construct the representations required by its original design. We do not modify the core architecture of an external baseline merely to enforce an artificial common input format. This evaluation practice is consistent with existing memory-based manipulation benchmarks, including RoboMME, RoboMemArena, and RMBench~\citep{dai2026robomme,lei2026robomemarena,chen2026rmbench}.

To complement this external comparison, we further introduce \mnm-VLA, which performs controlled studies within a shared framework. It isolates the effects of memory, state input, auxiliary supervision, high-level planning, and action-head design. This internal suite partly addresses the difficulty of directly comparing heterogeneous external methods and provides useful insights into individual design choices. This is also one of the main motivations for developing {\mnm}.

\paragraph{Is uniform sampling the optimal memory strategy?}Not necessarily. As discussed in the main paper, for perceptual memory, the key issue is not simply memory sparsity or the writing interval, but the stability of the writing trigger. If an event-driven trigger occurs too early or too late, it may write irrelevant observations or miss informative ones. Such noisy memory can further interfere with subsequent reasoning and lead to additional instability.

The chunk-based writing strategy avoids this problem by using a stable writing schedule. However, memory management is inherently more complex and also involves forgetting strategies and trade-offs between memory capacity and computational cost. We do not claim that the strategies explored in this work are optimal. Rather, we provide a framework for controlled comparison, and the evaluated strategies are sufficient to support the conclusions drawn in this paper.

Related studies in video understanding, such as FOCUS and AKS~\citep{zhu2025focusefficientkeyframeselection,tang2025adaptivekeyframesamplinglong},have also shown that uniform sampling often incurs only a limited performance drop compared with frame-selection methods based on learned visual scores, particularly for short videos. In embodied tasks, a video considered short in conventional video understanding may already correspond to a medium-horizon trajectory. This observation provides additional motivation for using regularly spaced sampling in manipulation tasks, where actions and scene contents often change gradually.

\paragraph{Can task success rate reflect active-perception capability?}
Yes. In {\mnm}, active perception is necessary for task completion: the robot must acquire missing information, retain it across views, and use it to guide subsequent actions. Therefore, task success rate provides a direct end-to-end measure of active-perception capability. Our failure decomposition further complements this metric by indicating whether failures mainly arise from missing relevant evidence, losing or failing to use context, or downstream reasoning and execution errors.
\section{Limitations and Future Work}
\label{app:limitations}

The current version of {\mnm} primarily supports the
Astribot S1. Future releases will extend
{\mnm} to additional embodiments, such as the Unitree G1
and AgiBot G2. We will also expand the benchmark
with more diverse and challenging tasks involving richer
interactions, longer perception--action horizons, and more
complex requirements for information acquisition, memory, and
closed-loop reasoning.

We plan to open-source the simulator, benchmark, datasets,
annotations, and evaluation tools, and continuously improve the
platform based on feedback from the research community. In
addition, we will provide more complete and standardized support
for existing and emerging robotic foundation models, evaluate a
broader range of VLA, memory-based, and world-action models,
and maintain a reproducible leaderboard. Our long-term goal is
to develop {\mnm} into an extensible and influential
benchmark for active perception and robotic manipulation.

\clearpage
\section{Additional Demonstrations}
\label{app:demos}
\subsection{Simulation Task Demonstrations}
See Fig.~\ref{fig:sim_vis_1},~\ref{fig:sim_vis_2},~\ref{fig:sim_vis_3},~\ref{fig:sim_vis_4},~\ref{fig:sim_vis_5},~\ref{fig:sim_vis_6},~\ref{fig:sim_vis_7},~\ref{fig:sim_vis_8},~\ref{fig:sim_vis_9}.

\begin{center}
\begin{minipage}{\linewidth}
\centering
\captionsetup{type=figure}
\includegraphics[width=\linewidth,height=0.70\textheight,keepaspectratio]{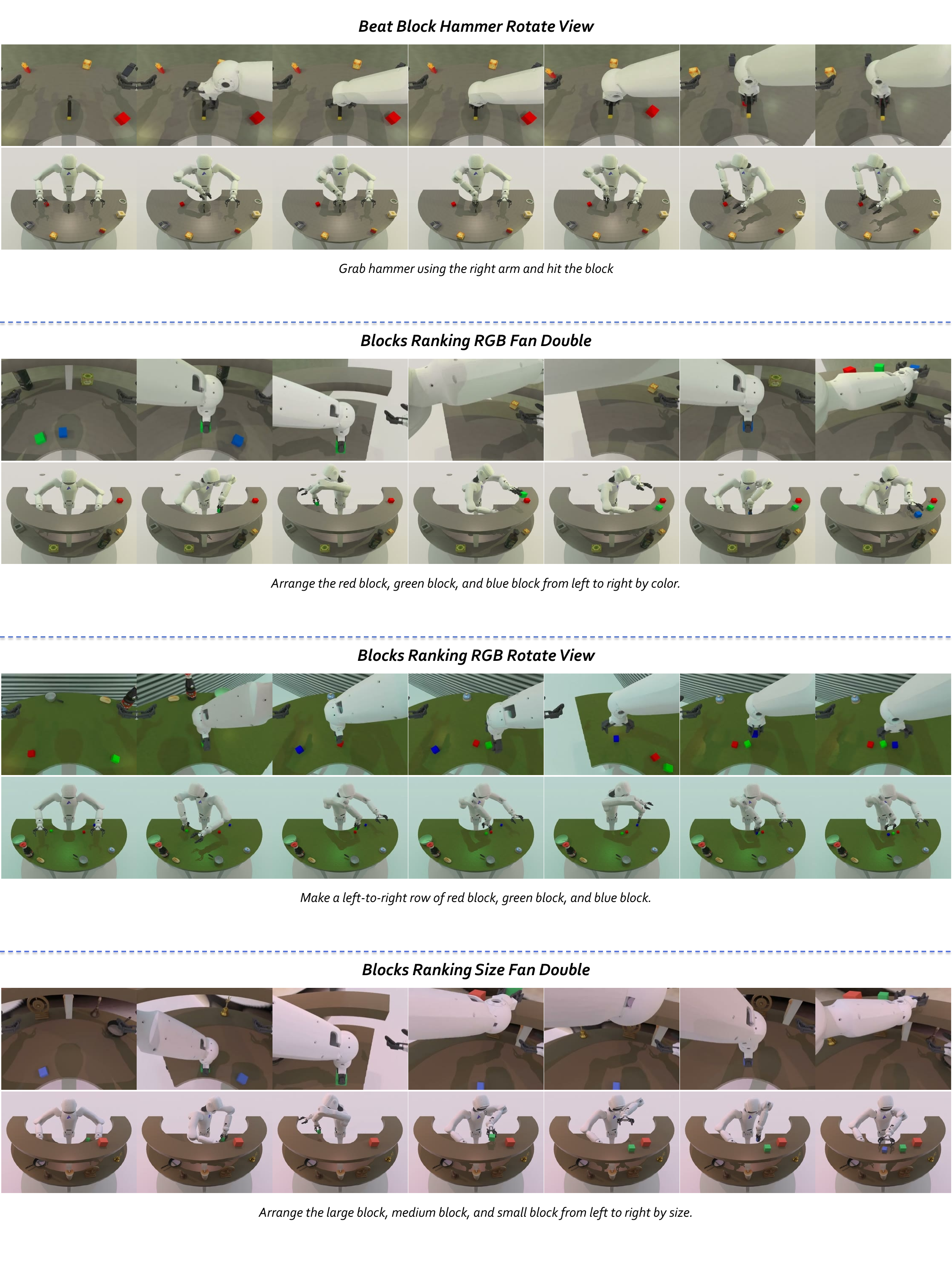}
\caption{Visualization of our custom simulator with all task.}
\label{fig:sim_vis_1}
\end{minipage}
\end{center}

\clearpage

\begin{center}
\begin{minipage}{\linewidth}
\centering
\captionsetup{type=figure}
\includegraphics[width=\linewidth,height=0.89\textheight,keepaspectratio]{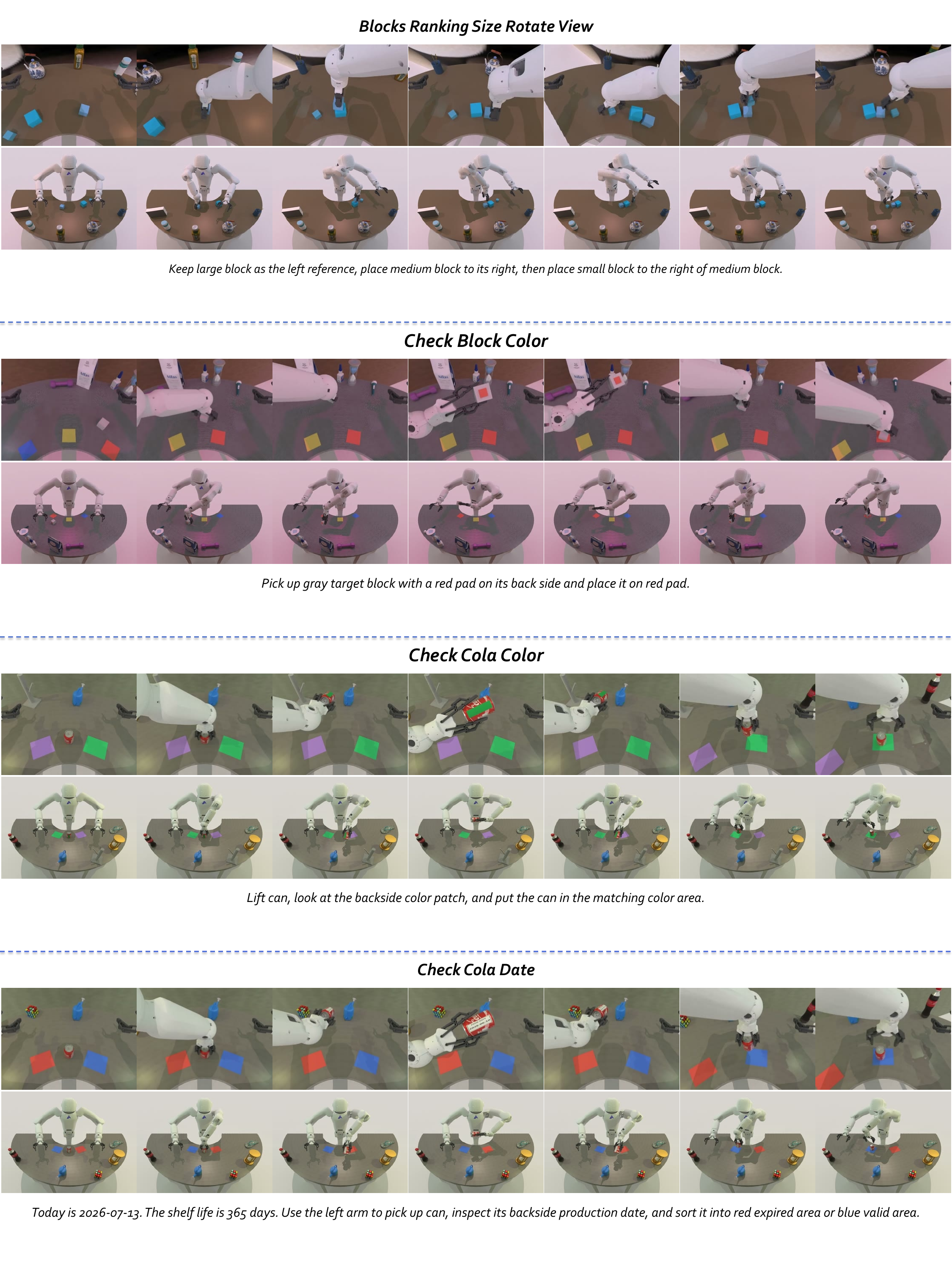}
\caption{Visualization of our custom simulator with all task.}
\label{fig:sim_vis_2}
\end{minipage}
\end{center}

\clearpage

\begin{center}
\begin{minipage}{\linewidth}
\centering
\captionsetup{type=figure}
\includegraphics[width=\linewidth,height=0.89\textheight,keepaspectratio]{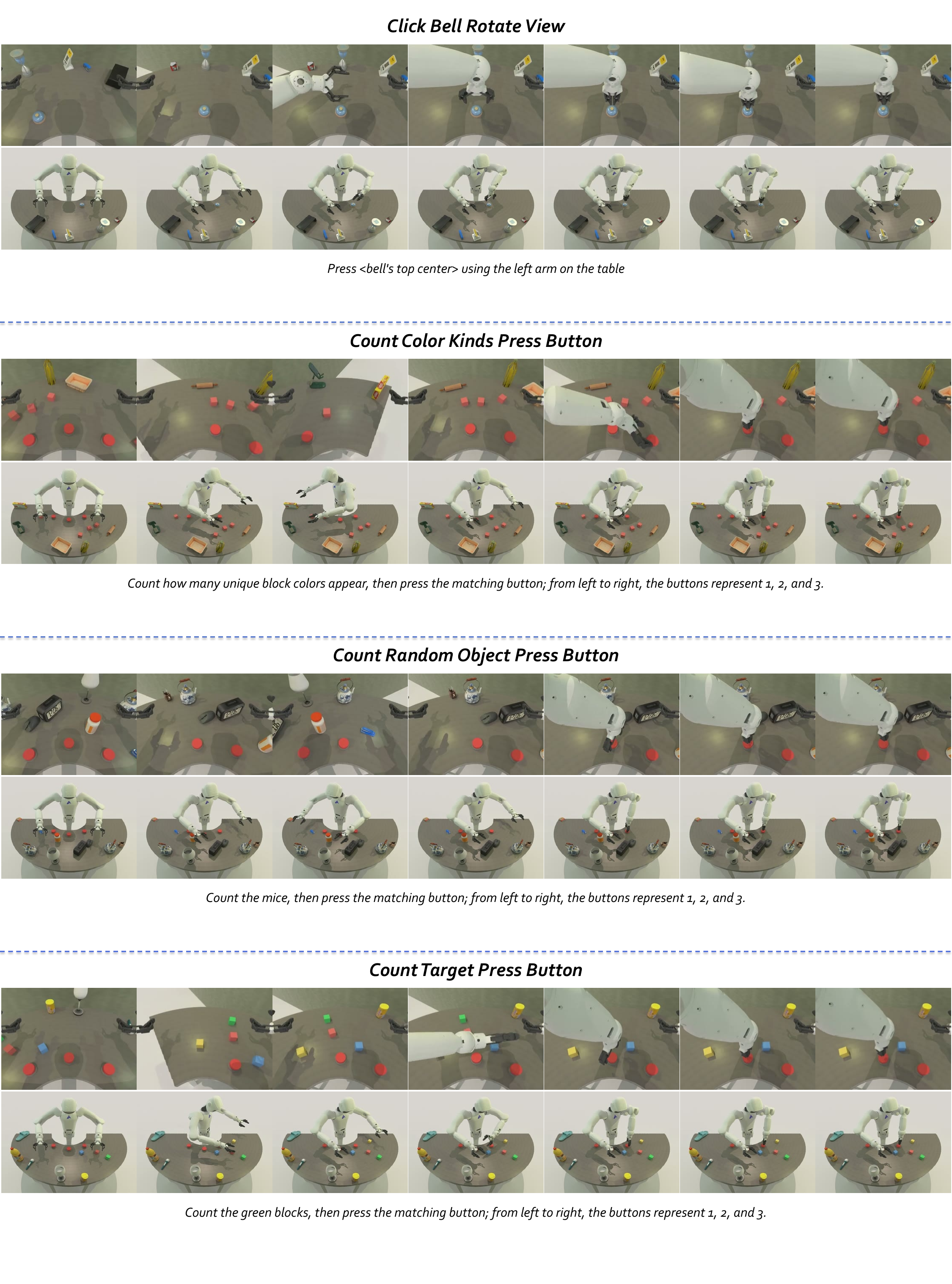}
\caption{Visualization of our custom simulator with all task.}
\label{fig:sim_vis_3}
\end{minipage}
\end{center}

\clearpage

\begin{center}
\begin{minipage}{\linewidth}
\centering
\captionsetup{type=figure}
\includegraphics[width=\linewidth,height=0.89\textheight,keepaspectratio]{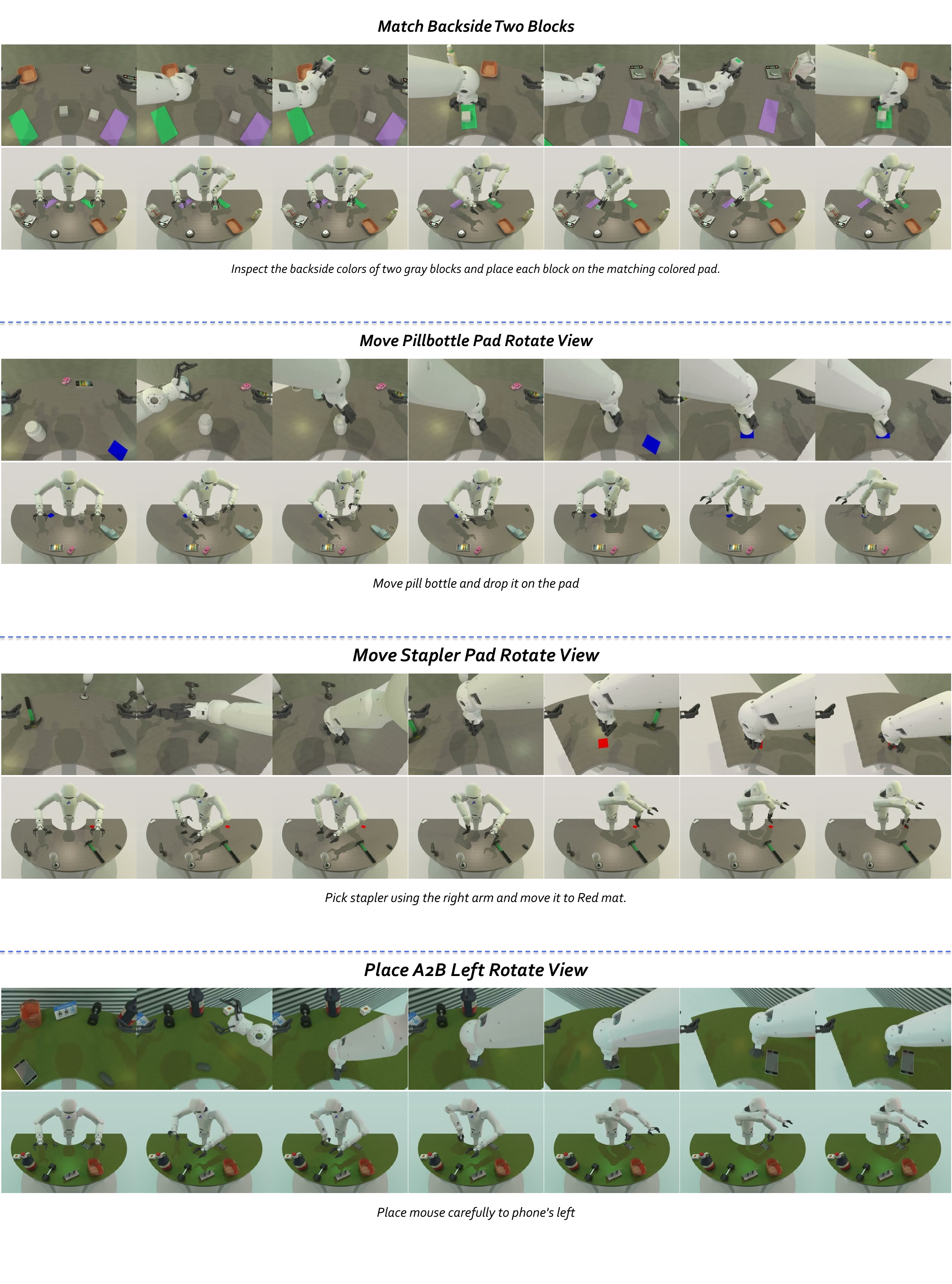}
\caption{Visualization of our custom simulator with all task.}
\label{fig:sim_vis_4}
\end{minipage}
\end{center}

\clearpage

\begin{center}
\begin{minipage}{\linewidth}
\centering
\captionsetup{type=figure}
\includegraphics[width=\linewidth,height=0.89\textheight,keepaspectratio]{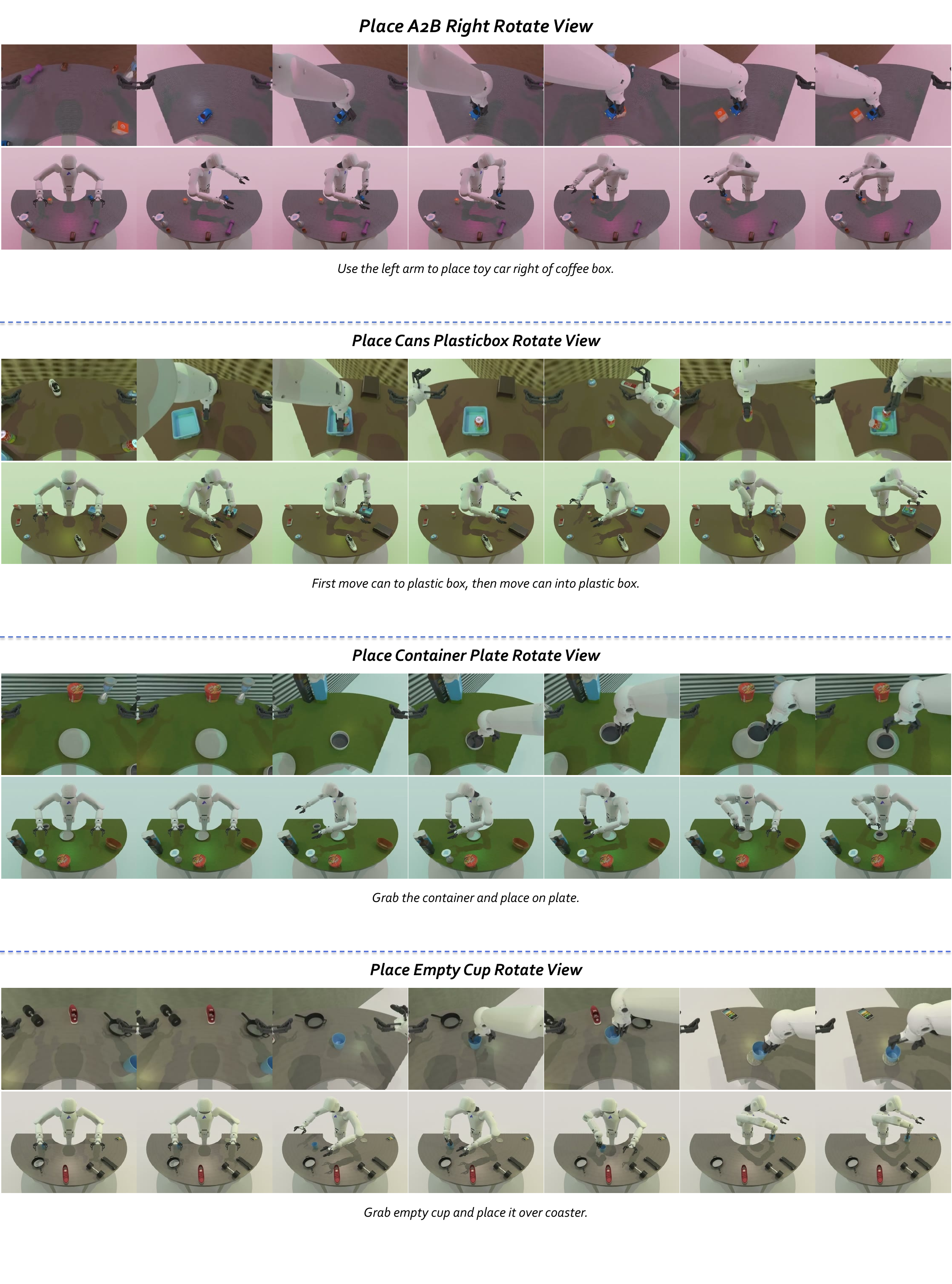}
\caption{Visualization of our custom simulator with all task.}
\label{fig:sim_vis_5}
\end{minipage}
\end{center}

\clearpage

\begin{center}
\begin{minipage}{\linewidth}
\centering
\captionsetup{type=figure}
\includegraphics[width=\linewidth,height=0.89\textheight,keepaspectratio]{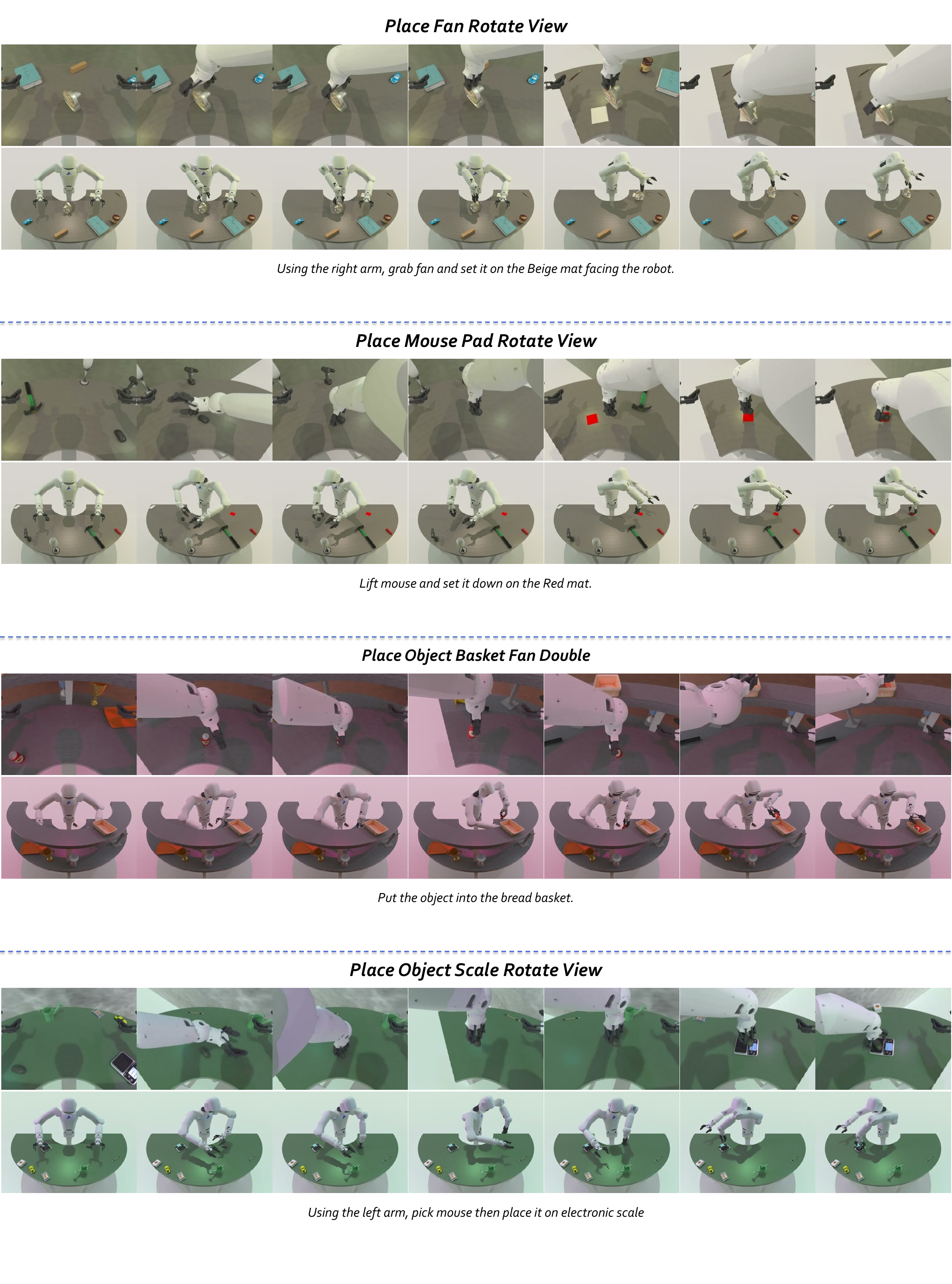}
\caption{Visualization of our custom simulator with all task.}
\label{fig:sim_vis_6}
\end{minipage}
\end{center}

\clearpage

\begin{center}
\begin{minipage}{\linewidth}
\centering
\captionsetup{type=figure}
\includegraphics[width=\linewidth,height=0.89\textheight,keepaspectratio]{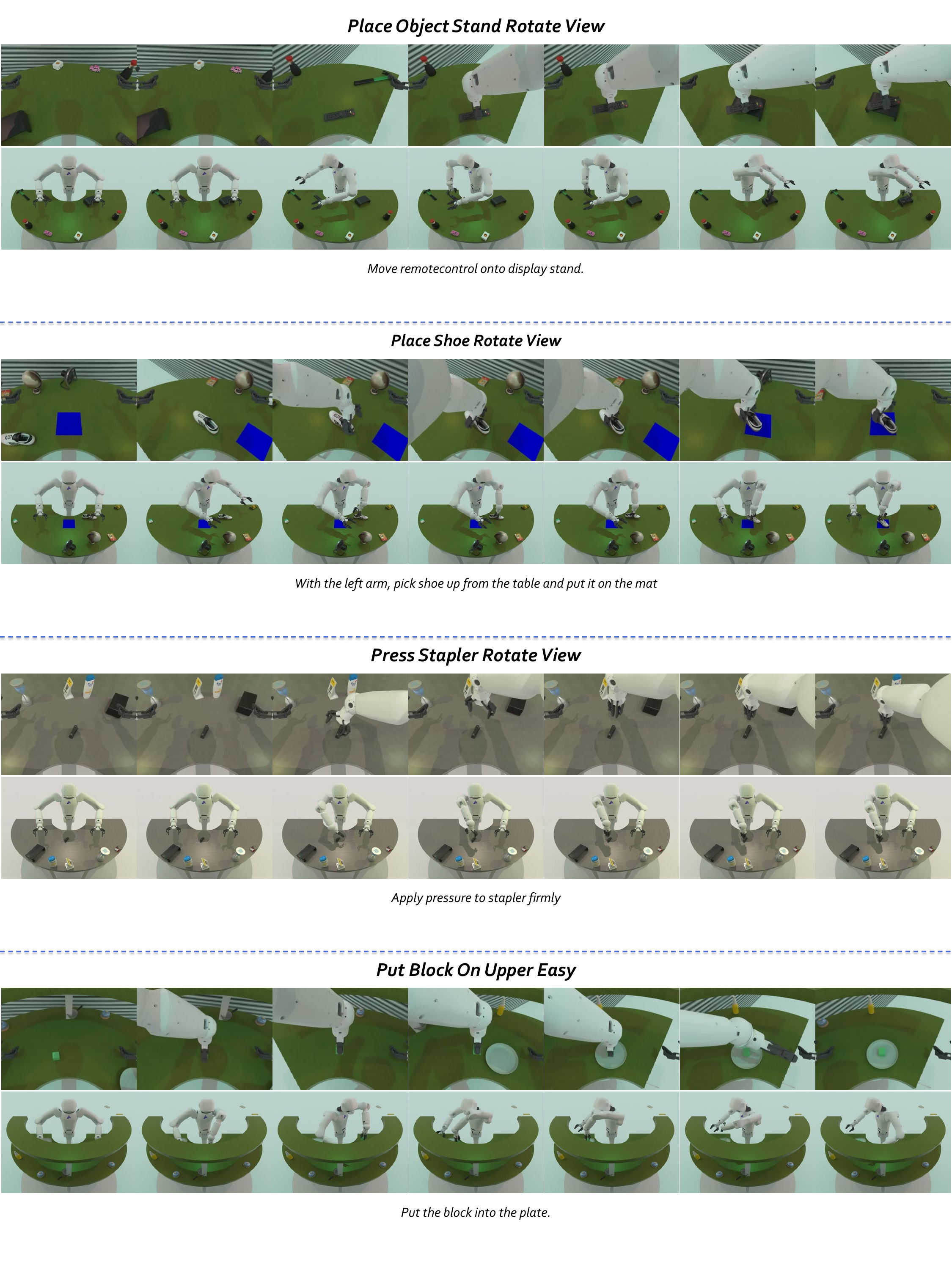}
\caption{Visualization of our custom simulator with all task.}
\label{fig:sim_vis_7}
\end{minipage}
\end{center}

\clearpage

\begin{center}
\begin{minipage}{\linewidth}
\centering
\captionsetup{type=figure}
\includegraphics[width=\linewidth,height=0.89\textheight,keepaspectratio]{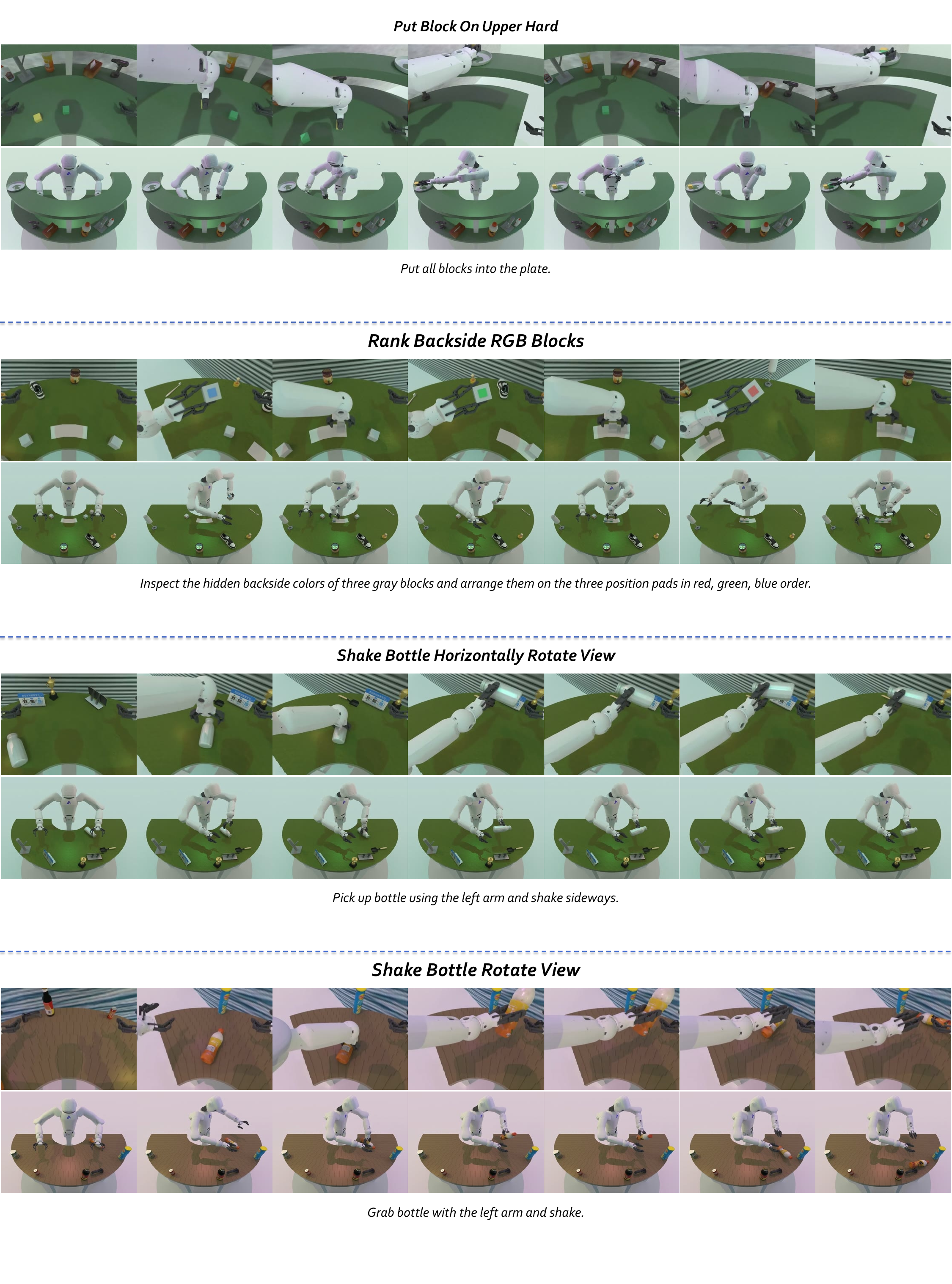}
\caption{Visualization of our custom simulator with all task.}
\label{fig:sim_vis_8}
\end{minipage}
\end{center}

\clearpage

\begin{center}
\begin{minipage}{\linewidth}
\centering
\captionsetup{type=figure}
\includegraphics[width=\linewidth,height=0.89\textheight,keepaspectratio]{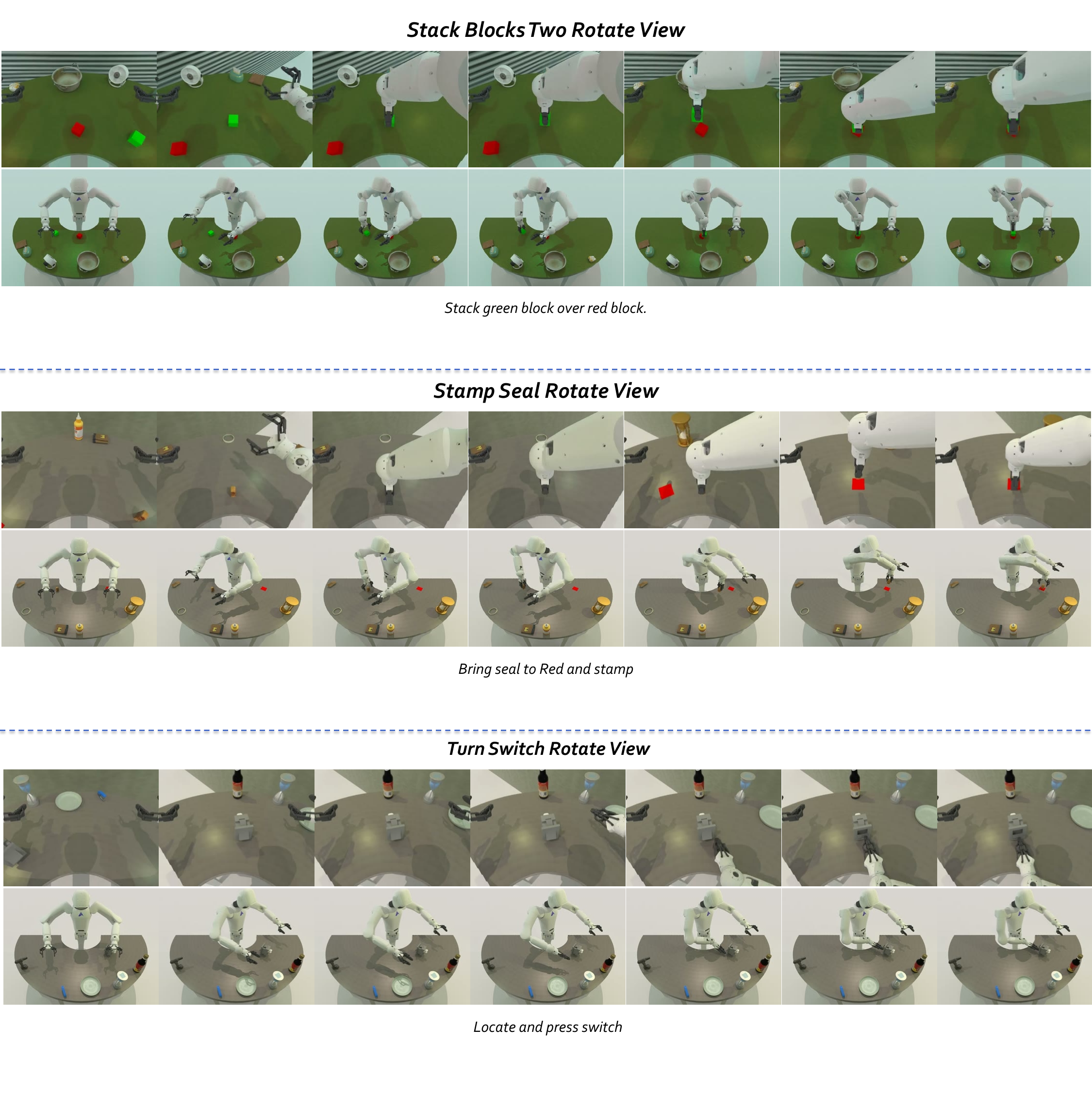}
\caption{Visualization of our custom simulator with all task.}
\label{fig:sim_vis_9}
\end{minipage}
\end{center}

\clearpage
\subsection{Real-World Task Demonstrations}
See Fig.~\ref{fig:real_vis}.
\begin{center}
\begin{minipage}{\linewidth}
\centering
\captionsetup{type=figure}
\includegraphics[width=\linewidth,height=0.79\textheight,keepaspectratio]{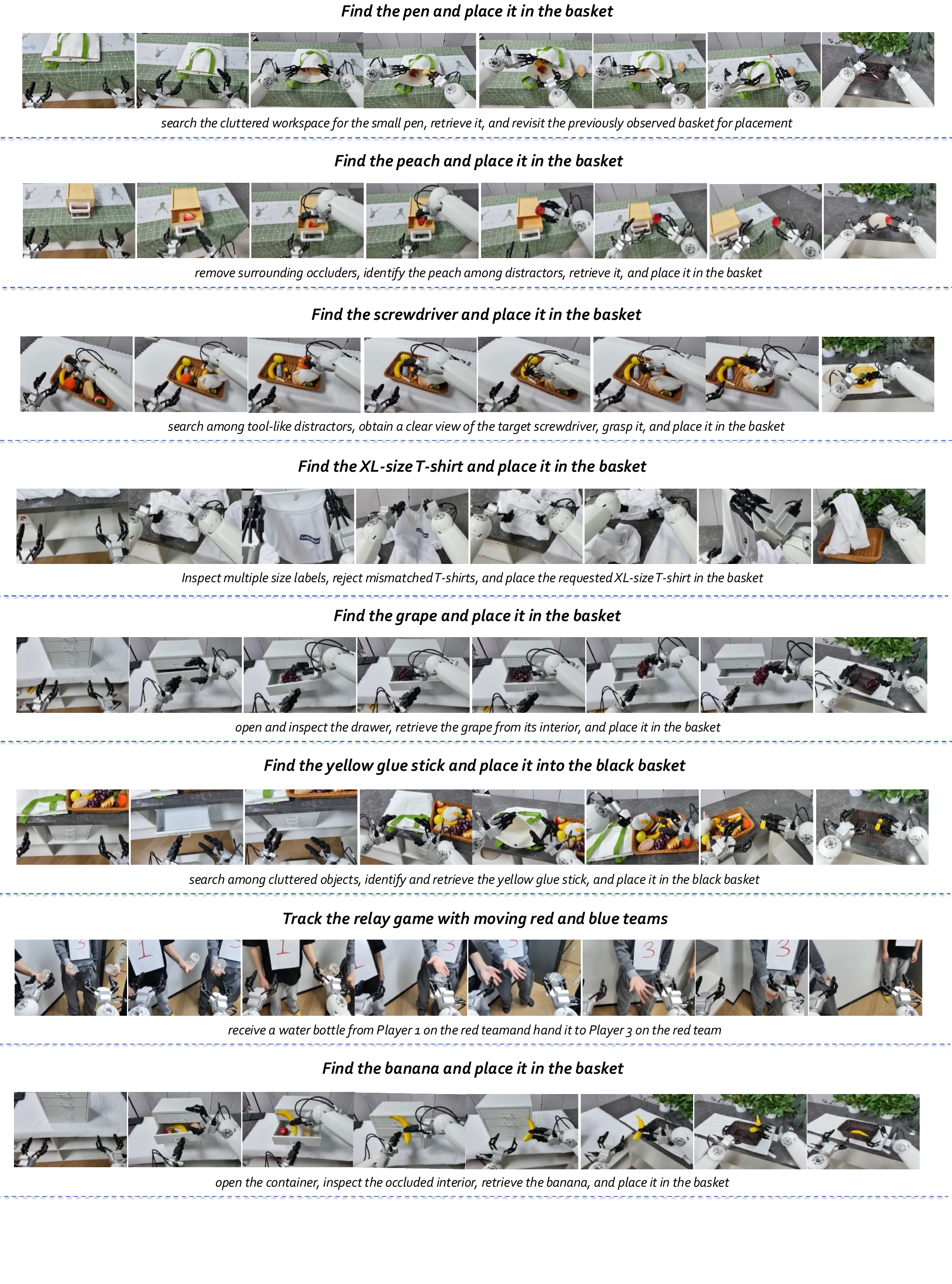}
\caption{Visualization of the eight real-world tasks.}
\label{fig:real_vis}
\end{minipage}
\end{center}

\end{document}